\documentclass[]{fairmeta}
\usepackage{float}
\title{
\sysname: Test-Time Training as Persistent Memory for Real-Time Video World Models
}

\author[1,2,*]{Junyi Wu}
\author[1,3,*]{Zekun Li}
\author[1]{Chenhongyi Yang}
\author[1]{Chen Guo}
\author[1]{Lingchen Yang} 
\author[1]{Yongan Zhang}
\author[1]{Xinyi Zhang}
\author[1]{Hongbing Hu}
\author[1]{Bartek Rymkowski}
\author[3,\dagger]{Srinath Sridhar}
\author[2,\dagger]{Yan Yan}
\author[1,\dagger,\ddagger]{Xijie Huang}

\affiliation[1]{Meta Reality Labs}
\affiliation[2]{University of Illinois Chicago}
\affiliation[3]{Brown University}

\contribution[*]{Equal contribution}
\contribution[\dagger]{Joint last author}
\contribution[\ddagger]{Project Lead}
\renewcommand\authorformat[2][]{\mbox{\small\sffamily\bfseries #2$^{#1}$}}
\usepackage[utf8]{inputenc}
\usepackage{url}
\usepackage{booktabs}
\usepackage{soul}
\usepackage{amsfonts}
\usepackage{amssymb}
\usepackage{amsmath}
\usepackage{mathtools}
\usepackage{nicefrac}
\usepackage{xcolor}
\usepackage{graphicx}
\usepackage{wrapfig}
\usepackage{enumitem}
\usepackage{tikz}
\usepackage{xspace}
\usepackage{times}
\usepackage{multirow}
\usepackage{colortbl}
\usepackage{wasysym}
\usepackage{array}
\usepackage{authblk}
\hypersetup{
  colorlinks=true,
  linkcolor=blue!60!black,
  citecolor=blue!60!black,
  urlcolor=blue!60!black
}

\newcommand{\sysname}{\mbox{World-T\textsuperscript{3}}\xspace}

\newcommand{\rlsttt}{Test-Time Memory Consolidation\xspace}

\newcommand{\rtdeploy}{Pipelined Streaming Rollout\xspace}

\newcommand{\kk}{\ensuremath{\mathbf{k}}}
\newcommand{\vvv}{\ensuremath{\mathbf{v}}}
\newcommand{\PP}{\ensuremath{\mathbf{P}}}
\newcommand{\WW}{\ensuremath{\mathbf{W}}}
\newcommand{\bphi}{\ensuremath{\boldsymbol{\phi}}}

\newcommand{\bPi}{\ensuremath{\boldsymbol{\Pi}}}

\abstract{
Recent advances in video world models have enabled plausible interactive environments, yet long-horizon generation places a fundamental demand on visual memory: the previously observed world must remain consistent as the model continues to generate. 
Existing autoregressive approaches often implicitly rely on the generative model itself to retain such information, leading to spatial memory decay beyond the active context.
We introduce \sysname, a camera-controlled video world model with a test-time training memory that consolidates past observations into dedicated model parameters. 
To robustly update this memory, we propose a recursive least-squares rule that incrementally absorbs novel visual features via online pose-keyed memory updates.
Our TTT memory reads out previously observed scene features from the fast-weight state by camera pose, enabling consistent generation with constant storage and per-step computation regardless of rollout length.
To facilitate real-time interactive generation with this robust memory, we further distill the autoregressive model into a few-step generator via dual-teacher distribution matching.
On a long-horizon revisit benchmark, \sysname raises revisit PSNR from $13.52$ to $18.31$\,dB and reduces absolute rotation error from $3.344^\circ$ to $1.075^\circ$ relative to the strongest baseline on both metrics, while sustaining $20.6$ FPS at $1280\times704$ on a single NVIDIA B200 GPU.
}

\date{\today}
\correspondence{\email{huangxijie1108@meta.com}}

\metadata[Project Page]{\url{https://adreamwu.github.io/World-T3/}}
\begin{document}

\maketitle

\begin{figure*}[h!]
%\vspace{-3mm}
\centering
\begin{subfigure}[b]{0.68\linewidth}
  \centering
  \includegraphics[width=\linewidth]{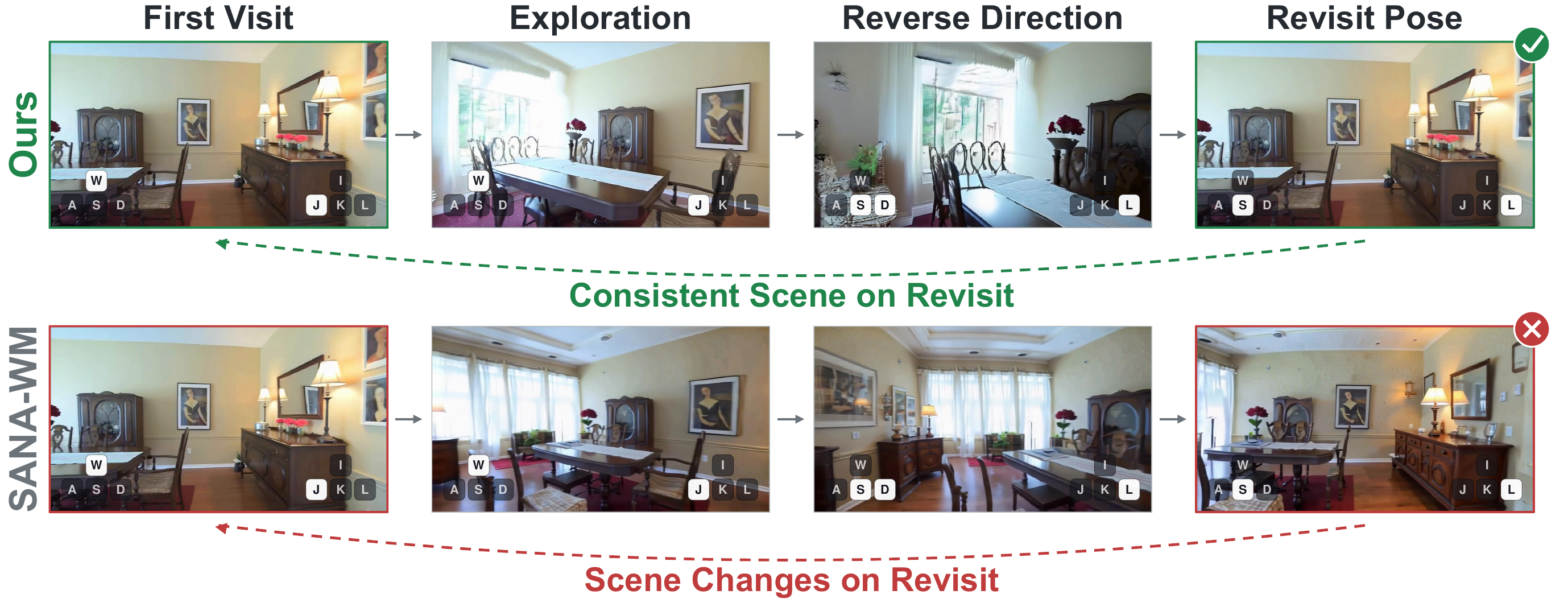}
  \phantomsubcaption\label{fig:teaser-demo}
\end{subfigure}\hfill
\begin{subfigure}[b]{0.31\linewidth}
  \centering
  \includegraphics[width=\linewidth]{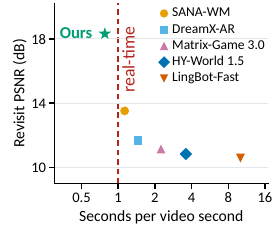}
  \phantomsubcaption\label{fig:teaser-speed}
\end{subfigure}
\vspace{-4mm}
\caption{\textbf{Left:} Revisit consistency.
When returning to a previously visited viewpoint, our \sysname enables faithful scene recall,
whereas existing autoregressive baselines forget earlier observations beyond the context window and resynthesize new content.
\textbf{Right:} Speed vs.\ quality on one NVIDIA B200 GPU.
$x$-axis: wall-clock seconds required per second of generated $16$\,FPS $1280\times704$ video (points left of the dashed line are real-time).
$y$-axis: PSNR on the long-horizon revisit benchmark.}
\vspace{-2mm}
\label{fig:teaser}
\end{figure*}

\section{Introduction}
\label{sec:intro}
Video world models aim to simulate how a world evolves in response to user control, emerging as a powerful interface for embodied simulation, playable environment generation, and virtual exploration~\citep{unisim2024,robotworldsim2026,genie2024,gamegenx2024,matrixgame2025,thematrix2024,lyra2_2026}. 
Conditioned on a reference image and a scene description, recent work utilizes autoregressive video diffusion models (VDMs) to allow users to explore and interact with generated environments through sequential control inputs, such as 6-DoF camera poses and natural language instructions~\citep{lingbotworld2026,dreamxworld2026,sanawm2026,alayaworld2026,wonder2026}. 
However, as these models move from short-horizon generation toward sustained interaction, generating a plausible next observation is no longer sufficient. 
The autoregressive VDMs tend to remain locally coherent while gradually losing earlier scene information once it falls outside the active history context, causing previously established content to be resynthesized from the generative prior rather than preserved as persistent world state.

Existing approaches attempt to preserve scene information beyond the local generation window by maintaining explicit 3D scene representations~\citep{voyager2025,spatia2025,mirage2026}, retrieving historical frames via camera and geometric cues~\citep{vmem2025,contextasmemory2025,worldmem2025,ucm2026,matrixgame3_2026}, or compressing past features into compact memory tokens~\citep{memrope2026,echoinfinity2026,gimworld2026}. 
While these methods improve long-horizon consistency, maintaining an expanding collection of reconstructed geometry, historical frames, or tokens often introduces unbounded storage overhead and escalating attention complexity as rollouts progress. 
This motivates a complementary question: can a video world model consolidate successive observations into an online-updated, parametric memory that enables recovery of previous scene content, while keeping memory storage and per-step access costs independent of rollout length?

To this end, we propose \sysname, a camera-controlled video world model that consolidates observation history into a fixed-size, pose-indexed parametric memory via test-time training (TTT).
As generated video blocks leave the local context, their pose-content associations are written into dedicated fast-weight memory, which is subsequently queried by incoming camera poses to recover features of previously observed regions.
A critical challenge in online parametric updating is handling smooth camera trajectories, where highly correlated pose features naturally introduce spatial redundancy across consecutive frames. 
To ensure that each update selectively incorporates novel visual content without degrading previously stored information, we propose \rlsttt (\S\ref{sec:ttmc}), an online recursive least-squares update rule that incorporates the accumulated covariance of pose keys into each write.
By solving a regularized regression objective over past pose-content associations, our method preconditions incoming key features to isolate non-redundant spatial information. 
Our TTT memory storage and per-step update complexity remain independent of rollout length, allowing history to be sequentially consolidated and accessed within a fixed budget.

Specifically, starting from a pretrained video diffusion model~\citep{wan2025}, we develop \sysname through a progressive three-stage training pipeline that establishes camera control, learns persistent scene memory, and enables few-step autoregressive generation (\S\ref{sec:method}).
Training draws on a curated corpus of 908K camera-annotated clips from real-world footage and synthetic environments (\S\ref{sec:data}).
In Stage 1, we adapt the pretrained backbone into a bidirectional teacher conditioned on 6-DoF camera trajectories.
In Stage 2, we then convert it into a block-causal autoregressive model with bounded visual context and jointly train the backbone and memory module to consolidate past observations and retrieve them by camera pose.
In Stage 3, to achieve real-time interactive generation, we design a dual-teacher distribution matching distillation framework~\citep{consistency2023,dmd2024,causalforcingpp2026,causvid2025} that compresses the multi-step autoregressive model into a 4-step generator, leveraging both the bidirectional and memory-augmented teachers to balance local visual quality and long-horizon memory consistency. 
At deployment, our \sysname sustains $20.6$\,FPS generation at $1280\times704$ on a single NVIDIA B200 GPU with live online memory updates.

Our main contributions are: 
(i) We introduce \sysname, a camera-controlled video world model that incorporates a pose-indexed test-time training memory. Through an online recursive least-squares update rule, our model robustly consolidates historical observations into a parametric memory whose storage and per-step computation remain constant regardless of rollout length.
(ii) We design a progressive three-stage training framework that converts a pretrained video diffusion backbone into an autoregressive world model with TTT memory, and further turn it into a few-step generator via dual-teacher distribution matching distillation.
(iii) Against five recent video world models, the four-step \sysname ranks first on all six evaluated revisit-consistency measures and is the only evaluated model to exceed the $16$\,FPS playback rate for real-time generation on a single NVIDIA B200 GPU.

\section{Related Work}
\label{sec:related}

\textbf{Video World Models.}
Recent advances in video generation have given rise to interactive world models that condition future observations on camera trajectories or action sequences~\citep{lingbotworld2026,dreamxworld2026,sanawm2026,alayaworld2026,wonder2026,genie2024,gamegenx2024,matrixgame2025,unisim2024}. To incorporate camera control, various representations have been explored, including pose vectors, dense Plücker ray maps, and relative positional encodings that integrate the geometric signal directly into the attention mechanism~\citep{cameractrl2025,motionctrl2024,scope2026,ucpe2025}. While these approaches produce visually compelling results over short horizons, they suffer from progressive camera drift and temporal incoherence as the generation window extends~\citep{wonder2026,contextforcing2026,mind2026,li2026ms}. The fundamental issue is persistence: a revisited region should faithfully preserve the state it exhibited on first encounter, rather than being hallucinated anew from the model's learned prior~\citep{loopnav2025,wrbench2026}. A parallel line of studies tackles this desideratum, whether by reconstructing and reprojecting previously observed geometry~\citep{voyager2025,spatia2025,mirage2026}, retrieving earlier frames according to spatial coverage~\citep{vmem2025,contextasmemory2025,worldmem2025,ucm2026,matrixgame3_2026}, or distilling the ever-growing observation history into a fixed-size token buffer~\citep{memrope2026,echoinfinity2026,gimworld2026}. These strategies, however, face an inherent dilemma: structured memory grows in cost with rollout length, while fixed-capacity alternatives inevitably dilute earlier observations as new content is compressed. In contrast, \sysname absorbs the growing history into the network weights via test-time training, incurring constant memory cost regardless of sequence length while preserving previously encountered states within the expressive capacity of the parameter space.

\textbf{Test-Time Training as Memory.}
Test-time training casts persistent memory as an online associative-memory problem that continuously absorbs new observations into a fixed-size parametric state~\citep{titans2025}. Successive works have progressively coarsened the write granularity from a single gradient step per token~\citep{tttrnn2024}, to objectives amortised over windows~\citep{atlas2025}, to chunk-level updates sized to saturate modern accelerators~\citep{lact2025}. The same principle has since migrated to visual and embodied domains: video diffusion models adopt it for minute-long coherent generation~\citep{tttvideo2025}, 3D reconstruction systems recast recurrent updating as online learning over image streams~\citep{ttt3r2026}, and spatial understanding and robot control leverage it to maintain persistent state over long streams or to distill reusable memory from unlabelled demonstrations~\citep{spatialttt2026,robottt2026,wamttt2026}.
Throughout this progression, a shared assumption remains: the gradient-based write presumes successive keys are diverse enough for a local step to meaningfully separate them. When keys are nearly collinear along camera poses in a smooth trajectory, each update largely reiterates what the store already contains. \sysname addresses this redundancy by solving the minimiser of the accumulated objective, providing video world models with a memory faithfully preserving scene state across long camera trajectories.

\section{\sysname}
\label{sec:method}

\sysname is built on three successive stages from a pretrained text-image-conditioned video diffusion model (VDM). 
(i) \emph{Bidirectional SFT} endows VDM with camera control and adapts it to the domain of freely navigated scenes using bidirectional attention, establishing high-fidelity generative priors (\S\ref{sec:stage1}). 
(ii) \emph{Autoregressive Training} introduces block-causal attention to enable interactive generation, augmented with persistent test-time training (TTT) memory for scene consistency (\S\ref{sec:stage2}).
(iii) \emph{Few-Step Distillation} further distills denoising sampling to only four steps for real-time streaming, leveraging dual-teacher guidance to preserve short-term visual quality and long-term memory (\S\ref{sec:stage3}).

\subsection{Stage 1: Bidirectional Camera-Conditioned Teacher}
\label{sec:stage1}
\begingroup
\setlength{\intextsep}{0pt}
\setlength{\columnsep}{10pt}
\begin{wrapfigure}{r}{0.45\textwidth}
\centering
\vspace{-1mm}
\includegraphics[width=0.93\linewidth]{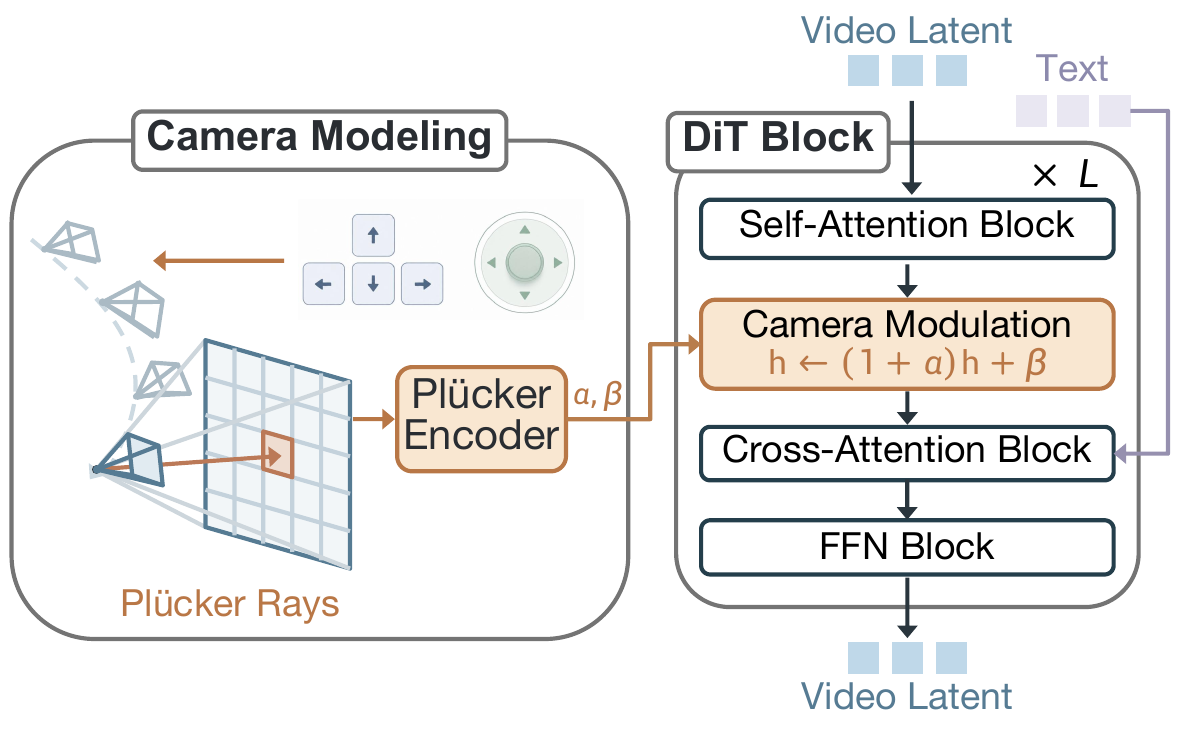}
\captionsetup{font=small,skip=2pt}
\caption{\textbf{Camera Conditioning.} Pl\"ucker raymaps modulate each DiT block via camera-dependent scale and shift.}
\label{fig:stage1_camera}
\end{wrapfigure}
To establish a strong foundation capable of producing globally coherent geometry and motion, Stage 1 adapts the pretrained VDM backbone with low-rank adapters on all attention and feed-forward projections, trained jointly with a newly introduced camera conditioning module, while the remaining backbone weights stay frozen. 
By attending over the entire temporal horizon simultaneously, the teacher constructs each latent frame with global awareness of all others, thereby realizing a given trajectory as a single coherent motion.
This bidirectional teacher model serves two pivotal roles in subsequent stages: it provides weight initialization for the autoregressive student model in Stage 2 (\S\ref{sec:stage2}), and supplies the reference score function for the distribution matching objective in Stage 3 (\S\ref{sec:stage3}).

We encode camera geometry using Pl\"ucker raymaps~\citep{cameractrl2025,scope2026}, with all poses expressed relative to the first frame. For a latent frame $f$ with rotation $\mathbf{R}_f$, camera origin $\mathbf{o}_f$, and intrinsics $\mathbf{K}$, the ray at spatial position $(u,v)$ has direction $\mathbf{d}_{f,uv} \propto \mathbf{R}_f\mathbf{K}^{-1}[u,v,1]^{\top}$ and is represented by its direction and moment:
$
\mathbf{r}_{f,uv} = \bigl(\mathbf{d}_{f,uv},\; \mathbf{o}_f \times \mathbf{d}_{f,uv}\bigr)
\in \mathbb{R}^{6}.
$

These rays form a six-channel field $\bPi$ aligned with the latent grid, providing spatially resolved camera conditioning. Identical viewpoints yield identical raymaps, which also serve as the basis for pose-keyed retrieval in our TTT memory module (\S\ref{sec:ttmc}). An embedding layer projects $\bPi$ into the VDM backbone's feature space and injects the features into every transformer block through adaptive layer-norm, with zero-initialized output projection for training stability.

\par
\endgroup

\subsection{Stage 2: Autoregressive Teacher with TTT Memory}
\label{sec:stage2}
Stage 1 endows the bidirectional backbone with camera control over fixed-length sequences. Despite strong generative priors, its non-causal nature precludes interactive and extended rollout.
In Stage~2, we convert this model into an autoregressive VDM via block-causal attention.
However, autoregressive generation over extended horizons inherently suffers from context decay, leading to severe visual hallucinations upon revisiting previous viewpoints.
To bridge this gap, we integrate a test-time training (TTT) memory into the autoregressive framework, preserving long-horizon scene consistency across rollout trajectories.

Specifically, generation produces a sequence of $N$ latent blocks $\mathbf{x}_{1:N} = (\mathbf{x}_1, \dots, \mathbf{x}_N)$, each containing $B$ latent frames. The rollout is conditioned on a text prompt $c$, a reference image $c_I$, and a 6-DoF camera trajectory $\mathcal{T}_{1:N}$. To keep per-block computation $O(1)$ \textit{w.r.t.} rollout length, we construct a bounded visual context $C_b$ and an evolving TTT memory state $\mathcal{M}_b$. The joint distribution factorizes block-causally as:
\begin{equation}
p_\theta(\mathbf{x}_{1:N} \mid \mathcal{T}_{1:N}, c_I, c)
= \prod_{b=1}^{N} p_\theta\!\left(
\mathbf{x}_b \mid C_b, \mathcal{M}_b, \mathcal{T}_{1:b}, c_I, c\right).
\end{equation}

\subsubsection{Block-Causal Adaptation}
\label{sec:context}
To maintain global identity while ensuring local continuity, 
as shown in Figure \ref{fig:stage2_ttmc} (a),
we construct the bounded visual context $C_b$ for each generation step $b$ as
$
C_b = \bigl[\;\mathbf{s}\;;\;
\mathbf{x}_{b-K:b-1}\;\bigr],
$
where the first generated block $\mathbf{s} = \mathbf{x}_1$ acts as a persistent sink to anchor long-term visual appearance, and the sliding window of $K$ recent clean blocks provides local temporal continuity across block boundaries.

We fine-tune the Stage~1 model using masked teacher forcing~\citep{diffusionforcing2024,causalforcing2026,causalforcingpp2026}. A block-causal attention mask ensures that the target block attends only to $C_b$ and its own noisy latents, excluding future blocks to allow parallel teacher-forced training.
% When generation advances and blocks are evicted from the sliding window, their representations are absorbed into $\mathcal{M}_b$ via the jointly trained TTT memory module.

\subsubsection{\rlsttt}
\label{sec:ttmc}
Long-horizon scene generation with viewpoint revisits demands that the model consistently recovers previously observed visual content. To achieve this, we propose \rlsttt, a test-time training module that represents each generated block as a key-value association between its camera pose and visual content. As generation progresses, blocks evicted from the active context window are consolidated into a fast-weight memory state via closed-form online updates~\citep{tttrnn2024,titans2025,atlas2025,lact2025,tttvideo2025,ttt3r2026}. The updated memory reconciles new observations with scene history, enabling subsequent generation steps to query and retrieve long-term visual information using camera pose as the key.

\begin{figure}[t]
\centering
\includegraphics[width=\linewidth]{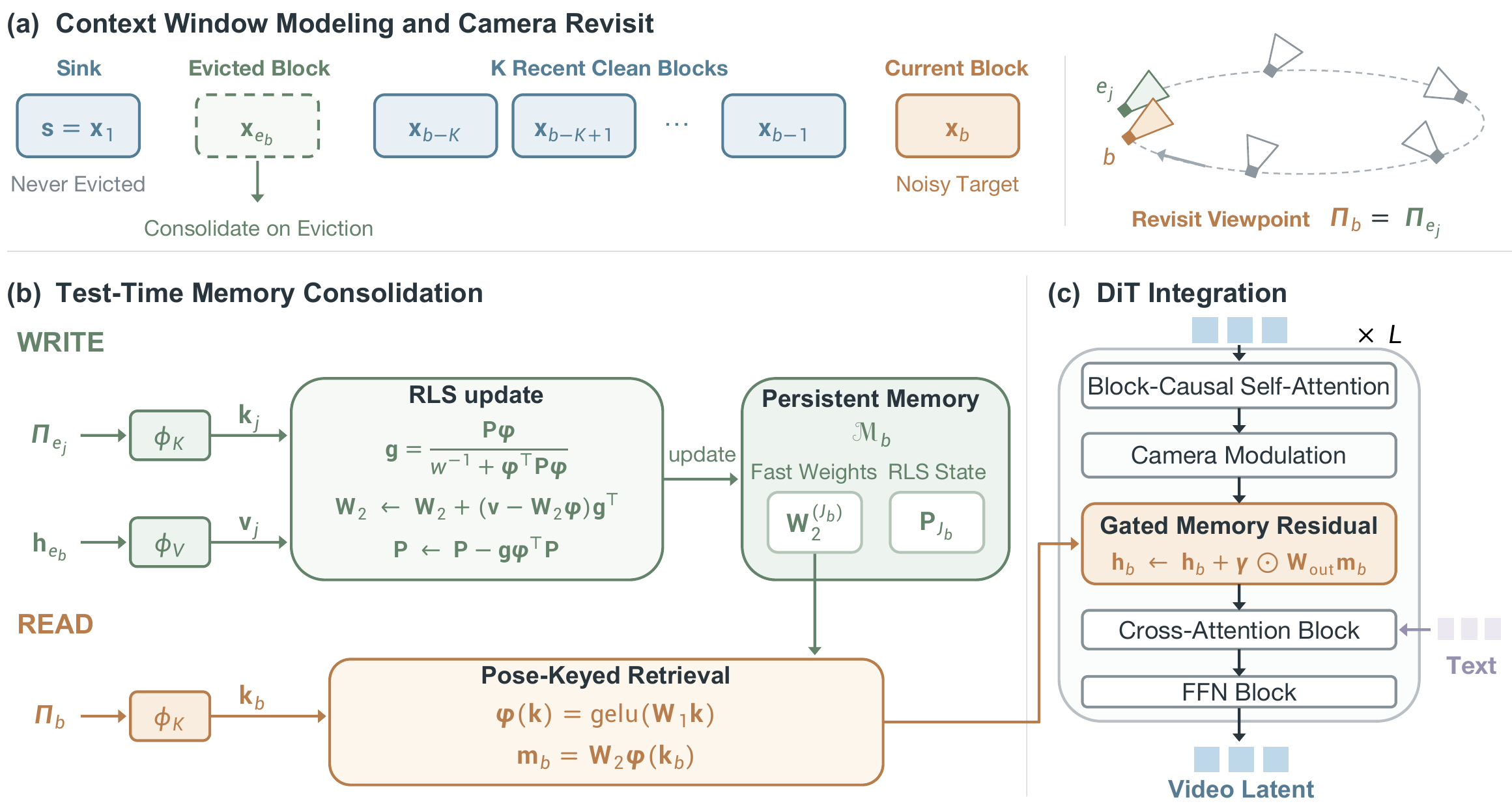}
%\vspace{-2mm}
\caption{\textbf{Stage 2: Autoregressive Teacher with TTT Memory.} \textbf{(a)} Overview of the context window modeling, which consists of a sink block, $K$ recent clean blocks, and the current block.
The revisit viewpoint may be evicted and thus fall outside the active context window.
\textbf{(b)} \textsc{Write} encodes an evicted block's raymap $\bPi_{e_j}$ and clean hidden state $\mathbf{h}_{e_j}$ into a key-value pair and updates TTT memory state $\mathcal{M}_b$ by recursive least squares (Eq. \ref{eq:rls}). \textsc{Read} uses the target pose key $\kk_b$ to retrieve memory feature $\mathbf{m}_b$ from the fast weights (Eq. \ref{eq:read}), without distorting the memory state with $J_b$ writes before block $b$.
\textbf{(c)} The retrieved memory is adaptively injected into each DiT block through a gated residual (Eq. \ref{eq:gate}) for long-horizon consistency during viewpoint revisits.
% consistent long-horizon world modeling.
}
\label{fig:stage2_ttmc}
%\vspace{-3mm}
\end{figure}

\textbf{Pose-Keyed Fast Weights.}
Each transformer layer in the backbone is augmented with a lightweight TTT module as parametric scene memory. The memory architecture consists of a two-layer MLP: a fixed projection layer $\WW_1$ that maps pose keys into a non-linear feature space via $\bphi(\mathbf{k}) = \mathrm{gelu}(\WW_1 \mathbf{k})$, and a dynamically updated readout matrix $\WW_2$ that stores accumulated scene representations across time. Crucially, memory operations are decoupled from diffusion noise, as keys depend solely on camera raymaps, while values are derived from clean latent representations.

\textbf{Consolidating Writes via Recursive Least Squares.}
When block $e_j$ is evicted from the context window at the $j$-th write step, its camera raymap $\bPi_{e_j}$ and hidden state $\mathbf{h}_{e_j}$ are projected into a key-value association:
\begin{equation}
\kk_j = \varphi_K(\bPi_{e_j}), \qquad \vvv_j = \varphi_V(\mathbf{h}_{e_j}),
\label{eq:kv_encode}
\end{equation}
where $\varphi_K$ and $\varphi_V$ denote spatial pose and content encoders, respectively.
As illustrated in Figure \ref{fig:stage2_ttmc} (b),
upon context eviction, the pair $(\kk_j, \vvv_j)$ is consolidated into the TTT memory state $\mathcal{M}_b$.

Memory updates are conducted in an online manner, with key-value blocks presented sequentially upon context eviction. However, when a camera traverses a scene smoothly, consecutive evictions often originate from closely aligned viewpoints, resulting in correlated and near-collinear pose keys. A naive write strategy would repeatedly reinforce these dominant directions, allowing redundant observations to bias the stored memory. To ensure each write contributes novel information, we formulate memory consolidation as a weighted ridge regression problem over the key-value pairs observed up to write $j$:
\begin{equation}
\min_{\WW_2}\; \sum_{i \le j} w_i \bigl\lVert \vvv_i - \WW_2\bphi_i \bigr\rVert^{2}
\;+\; \lambda \bigl\lVert \WW_2 - \WW_2^{(0)} \bigr\rVert_F^{2},
\label{eq:ridge}
% \vspace{-2mm}
\end{equation}
where $\bphi_j = \mathrm{gelu}(\WW_1\kk_j)$ is the pose key feature, $w_j$ is a learned scalar weight, and $\WW_2^{(0)}$ is the learned base readout matrix.
To solve Eq.~\ref{eq:ridge} online per eviction, we recursively track the inverse regularized feature correlation matrix $\PP_j = \bigl(\sum_{i \le j} w_i \bphi_i \bphi_i^{\!\top} + \lambda\mathbf{I}\bigr)^{-1}$. Starting from pre-write states $\WW_2 = \WW_2^{(j-1)}$ and $\PP = \PP_{j-1}$, the recursive least squares update rule for the $j$-th write is:
\begin{equation}
\mathbf{g} = \frac{\PP\bphi}{w^{-1} + \bphi^{\!\top}\PP\bphi},
\qquad
\WW_2 \leftarrow \WW_2 + \bigl(\vvv - \WW_2\bphi\bigr)\mathbf{g}^{\!\top},
\qquad
\PP \leftarrow \PP - \mathbf{g}\,\bphi^{\!\top}\PP ,
\label{eq:rls}
\end{equation}
where index $j$ is omitted for brevity. The update direction $\PP\bphi$ acts as a whitening preconditioner that attenuates feature components along directions already spanned by prior memory, ensuring that only the novel residual updates $\WW_2$. Since $\WW_1$ remains fixed, Eq.~\ref{eq:rls} yields the exact minimizer of Eq.~\ref{eq:ridge} at every step, and therefore the consolidated memory state is invariant to the observation order.

We apply Eq.~\ref{eq:rls} once per eviction step per layer and yield the memory state $\mathcal{M}_b = \bigl(\WW_2^{(J_b)},\, \PP_{J_b}\bigr)$ for generating target block $b$, where $J_b$ counts the total writes completed prior to block $b$. We set $\lambda = 10^{-3}$ and initialize $\PP_0 = \mathbf{I}/\lambda$. This incorporates the base prior $\WW_2^{(0)}$ while stabilizing the whitening process to prevent unbounded amplification of subtle key residuals.

\textbf{Memory Retrieval and Injection.}
During generation of a target block $b$, memory reading operates using the target camera raymap $\bPi_b$ via its pose key $\kk_b = \varphi_K(\bPi_b)$.
As shown in Figure \ref{fig:stage2_ttmc} (b),
memory retrieval is performed via a forward pass:
\begin{equation}
\mathbf{m}_b = \mathrm{read}(\kk_b) = \WW_2 \, \bphi(\kk_b) = \WW_2 \, \mathrm{gelu}(\WW_1 \kk_b).
\label{eq:read}
\end{equation}
The retrieved memory feature $\mathbf{m}_b$ is then projected back to the backbone feature dimension and injected into the layer's hidden state $\mathbf{h}_b$ via a learned per-channel gate $\boldsymbol{\gamma}$ as shown in Figure \ref{fig:stage2_ttmc} (c):
\begin{equation}
\mathbf{h}_b \leftarrow \mathbf{h}_b + \boldsymbol{\gamma} \odot \mathbf{W}_{\text{out}} \, \mathbf{m}_b .
\label{eq:gate}
\end{equation}
\textbf{Outer-Loop Learning.}
The TTT memory module is trained jointly with the DiT backbone. Gradients from the standard VDM objective propagate through the recursive update in Eq.~\ref{eq:rls} to optimize the key projector $\WW_1$, the base readout matrix $\WW_2^{(0)}$, the encoders $\varphi_K, \varphi_V$, the adaptive sample weights $w$, and the gating parameters $\boldsymbol{\gamma}$. By jointly optimizing the update mechanism, this outer-loop training aligns the TTT memory with the DiT backbone, enabling the model to dynamically adapt its internal state while preserving a shared representation space.
The distinction between learned parameters and online state is central to our formulation. During deployment, the backbone, pose and content encoders, and key projector remain fixed, and only the readout matrix and inverse feature-correlation state evolve through Eq~\ref{eq:ridge}. Each write therefore incorporates a clean observation without backpropagating through the diffusion transformer, while subsequent reads condition generation on the accumulated scene history.

\subsection{Stage 3: Dual-Teacher Few-Step Distillation}
\label{sec:stage3}
The autoregressive model in Stage~2 requires tens of sampling steps per block during generation, which is computationally incompatible with real-time interactive generation. Therefore, the goal of Stage~3 is to accelerate the generation process to a few steps while preserving both visual quality and long-term memory.

Following \citet{causalforcingpp2026}, we initialize the few-step student via Consistency Distillation, which discretizes the continuous flow-matching into a 4-step schedule.

\textbf{Dual-Teacher Distribution Matching Distillation.}
While consistency distillation enables fast sampling, direct multi-step jumps tend to degrade frame-level visual fidelity and spatial consistency across frames.
To recover both, we design a dual-teacher framework with Distribution Matching Distillation (DMD)~\citep{dmd2024, selfforcing2025}, which factorizes the real-data score into complementary guidance: 
\begin{itemize}
    \item \emph{Bidirectional Score} ($\mathbf{s}_{\text{bidir}}$), provided by the Stage 1 teacher operating over its fixed temporal window, enforcing short-range visual realism and motion coherence.
    \item \emph{Memory-Augmented Autoregressive Score} ($\mathbf{s}_{\text{mem}}$), provided by the Stage 2 teacher with TTT memory, evaluating latents along extended rollouts to preserve long-horizon spatial consistency.
\end{itemize}

To compute distribution matching gradients, each teacher is paired with an architecturally isomorphic critic network trained on student-generated samples. Specifically, $\text{critic}_{\text{bidir}}$ mirrors the bidirectional teacher for local quality, while $\text{critic}_{\text{mem}}$ mirrors the autoregressive teacher with an identical TTT memory module. The generator gradient is then formulated by distribution matching updates:
\begin{equation}
\nabla_\theta \mathcal{L}_{\mathrm{DMD}} = \nabla_\theta \mathcal{L}_{\mathrm{DMD}}\bigl(\text{teacher}_{\text{bidir}}, \text{critic}_{\text{bidir}}\bigr) + \nabla_\theta \mathcal{L}_{\mathrm{DMD}}\bigl(\text{teacher}_{\text{mem}}, \text{critic}_{\text{mem}}\bigr),
\label{eq:dual_dmd}
\end{equation}
where each term computes the score-difference gradient on student rollout latents by:
\begin{equation}
\nabla_\theta \mathcal{L}_{\mathrm{DMD}}(\text{teacher}, \text{critic}) = -\,\mathbb{E}_{t, \mathbf{z}_t}\Biggl[ \bigl( \hat{\mathbf{x}}_{0,\text{teacher}}(\mathbf{z}_t) - \hat{\mathbf{x}}_{0,\text{critic}}(\mathbf{z}_t) \bigr) \frac{\partial \mathbf{x}_\theta}{\partial \theta} \Biggr].
\label{eq:dmd_term}
\end{equation}
Here, $\mathbf{x}_\theta$ is the clean block generated by the few-step student under parameters $\theta$, and $\mathbf{z}_t$ is its noised counterpart at a randomly sampled timestep $t$.
Prior to computing $\mathbf{s}_{\text{mem}}$, the memory states of both the autoregressive teacher and its critic are initialized by running their memory update schedule over the student's rollout.
Thus, the memory-augmented guidance penalizes cumulative trajectory drift and inconsistent generation of the student under its own rollout conditions.

\section{Experiments}
\label{sec:experiments}
\vspace{-0.3cm}
In this section, we evaluate whether \sysname recovers previously observed scenes beyond its recent context window while following the requested camera trajectory. We measure revisit fidelity, camera control, and visual quality, then examine larger camera sweeps and 4-step distillation. Together, these experiments assess the two requirements motivating our design: retaining scene history within a bounded memory budget and supporting real-time generation.

\begin{figure}[H]
    \centering
    \includegraphics[width=\linewidth]{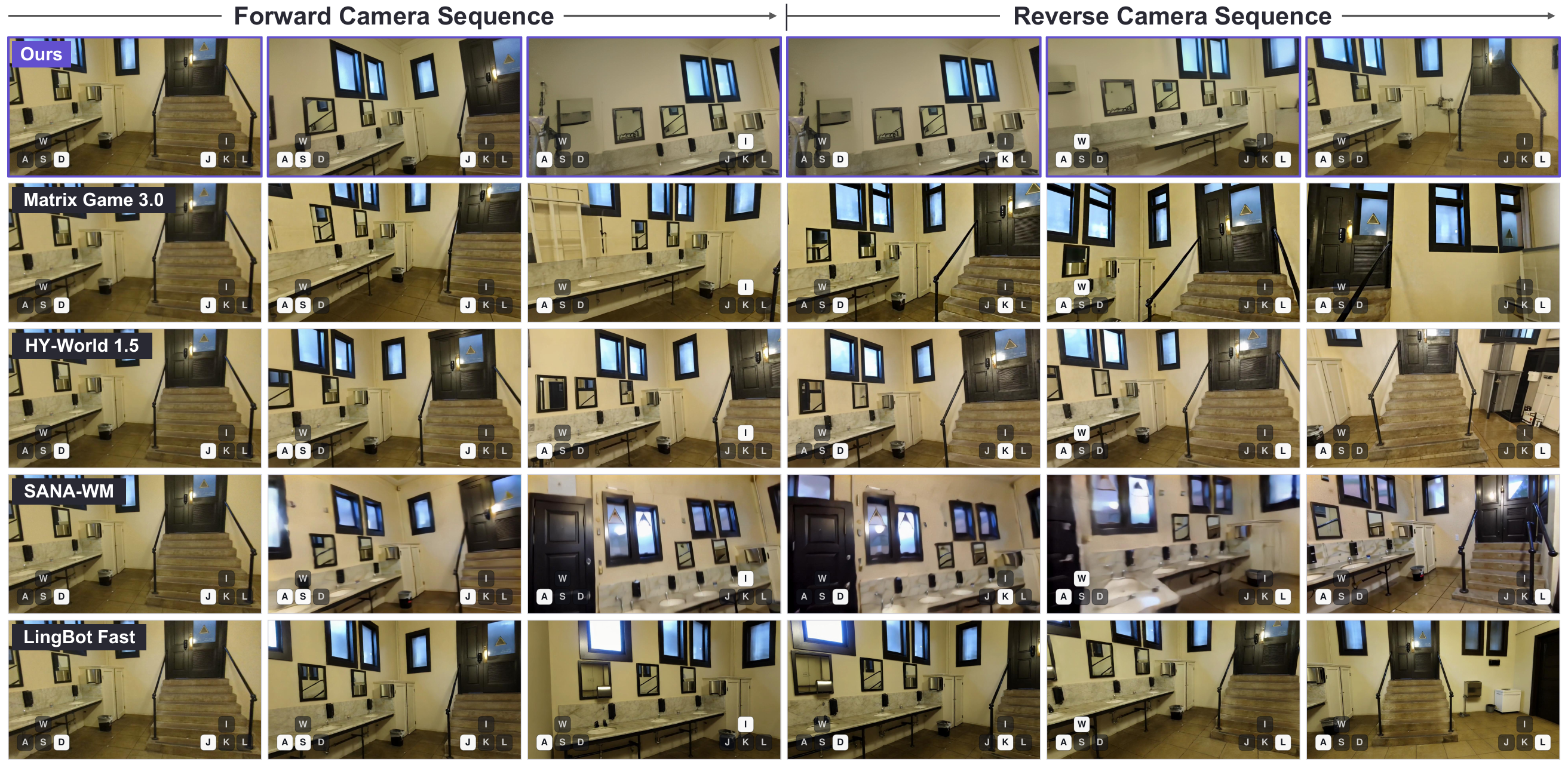}
    \par\vspace{2mm}
    \includegraphics[width=\linewidth]{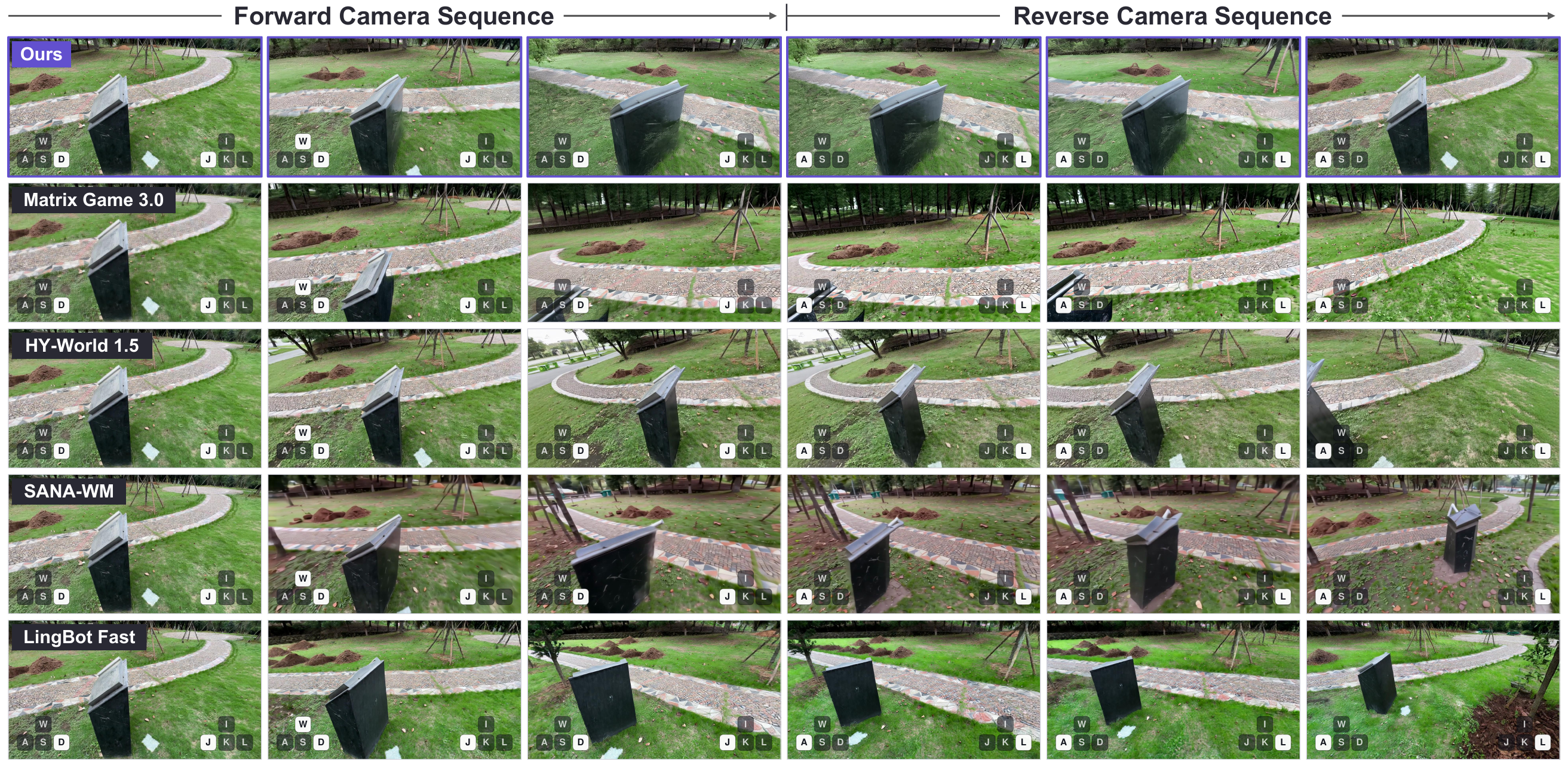}
    %\vspace{-4mm}
    \caption{\textbf{Scene Consistency along Out-and-Back Camera Trajectories.} Qualitative comparisons on an indoor scene (top) and an outdoor scene (bottom). Each 10-second camera input follows the forward trajectory for 5 seconds and then retraces it in reverse for 5 seconds. Columns show selected timestamps shared across methods, and the first and last columns compare the initial and return viewpoints.}
    \label{fig:revisit-memory}
    %\vspace{-3mm}
\end{figure}

\subsection{Evaluation Protocol}
\label{sec:setup}
\textbf{Revisit Benchmark.}
To rigorously test persistent scene memory upon viewpoint revisits, we construct a dedicated benchmark comprising $120$ clips with out-and-back camera trajectories sampled from the test set of our curated data.
Each sequence spans $160$ frames ($10$\,s at $16$\,FPS) at $1280\!\times\!704$ resolution, with the camera reversing its path at frame $80$. By design, these trajectories force the camera to re-observe the initial environment after a substantial temporal excursion. All evaluated methods are conditioned on identical initial frames, text prompts, and prescribed 6-DoF camera trajectories. Appendix~\ref{app:replay_protocol} provides more details on the curation and preprocessing.

\textbf{Baselines.}
We compare against Matrix-Game~3.0~\citep{matrixgame3_2026}, HY-World~1.5~\citep{worldplay2025}, DreamX-AR~\citep{dreamxworld2026}, LingBot-Fast~\citep{lingbotworld2026}, and SANA-WM~\citep{sanawm2026}, using their official checkpoints and code. We evaluate both the Stage~2 autoregressive teacher and the Stage~3 few-step student of \sysname.

\textbf{Revisit Fidelity.}
Adapting the revisit protocol of WorldMark~\citep{worldmark2026}, we quantify persistent memory retention by aggregating revisit consistency across four temporal gaps. Since the trajectory reverses at frame $80$, we pair outbound frames $t$ with return frames $160-t$ at the identical requested camera pose. Specifically, we evaluate the zero-indexed frame pairs $(8,152)$, $(16,144)$, $(24,136)$, and $(32,128)$, corresponding to temporal separations of $36$, $32$, $28$, and $24$ latent frames, respectively. Crucially, all four intervals exceed our $21$-latent-frame active context window. We average the DINOv2 cosine distances over these pairs and report one minus this mean distance as semantic similarity. To assess reference fidelity, we also compute the PSNR between each generated return view and its corresponding VAE-decoded ground-truth (GT) view, averaging across the four paired frames. We compare all views at their prescribed timestamps without post-hoc alignment, ensuring both metrics remain naturally sensitive to trajectory-following deviations. Further details are provided in Appendix~\ref{app:replay_metrics}.

\textbf{Camera Control and Visual Quality.}
We reconstruct camera trajectories from generated videos using VGGT-$\Omega$~\citep{vggtomega2026} and report absolute rotation error (RotErr), camera-matrix error (CamMC), alongside consecutive-frame rotational and translational relative pose errors (RPE). We align translations using a globally fitted scalar, preserving the unified initial-frame coordinate system. All camera-following metrics are macro-averaged across source datasets, with VAE-decoded GT serving as a reference for estimation error.
For visual quality, we report the average across seven VBench~\citep{vbench2024} dimensions: imaging quality, aesthetic quality, subject consistency, background consistency, motion smoothness, temporal flickering, and overall consistency. Detailed metric definitions and aggregation rules are provided in Appendix~\ref{app:replay_metrics}. Statistical significance in paired comparisons is assessed using exact two-sided sign tests, as elaborated in Appendix~\ref{app:paired_results}.

\begin{table*}[ht]
\centering
\small
\setlength{\tabcolsep}{4.pt}
\caption{\textbf{Comparison on the Revisit Set.} We average across seven VBench visual quality dimensions ($0$-$100$ scale). Revisit fidelity evaluates scene consistency via DINOv2 similarity (``Semantic Sim.'') between paired views and via PSNR against ground truth (GT).
``Mem'' is peak memory during generation on a single NVIDIA B200 GPU in GB.
Bold marks the best generated result, and underline indicates the second best. 
VAE-decoded GT provides a reference for the evaluation.}
% \vspace{-2mm}
\setlength{\tabcolsep}{1mm}
\resizebox{1\textwidth}{!}{%
\begin{tabular}{l c c cc cccc cc}
\toprule
& & \multicolumn{1}{c}{Visual Quality $\uparrow$} & \multicolumn{2}{c}{Revisit Fidelity $\uparrow$} & \multicolumn{4}{c}{Camera Following $\downarrow$} & \multicolumn{2}{c}{Efficiency}\\
\cmidrule(lr){3-3}\cmidrule(lr){4-5}\cmidrule(lr){6-9}\cmidrule(lr){10-11}
Method & Steps & VBench & Semantic Sim. & PSNR & RotErr$^\circ$ & CamMC & RPE Rot.$^\circ$ & RPE Trans. & FPS $\uparrow$ & Mem $\downarrow$  \\
\midrule
GT (VAE-decoded) & -- & 75.69 & 0.9982 & -- & 0.194 & 0.0055 & 0.073 & 0.00102 & -- & -- \\
\midrule
Matrix-Game 3.0 & 3 & 74.72 & 0.8225 & 11.15 & 11.179 & 0.2958 & 0.357 & 0.00285 & 7.1 & 76 \\
HY-World 1.5 & 4 & \textbf{75.32} & 0.8563 & 10.84 & 12.864 & 0.3484 & 0.315 & 0.00395 & 4.5 & 110 \\
DreamX-AR & 4 & 74.83 & 0.8768 & 11.66 & 8.011 & 0.2276 & 0.448 & 0.00508 & 11.0 & 76 \\
LingBot-Fast & 4 & 74.98 & 0.8580 & 10.59 & 9.817 & 0.2639 & 0.317 & 0.00290 & 1.6 & 144 \\
SANA-WM & 4+3\textsuperscript{$\dagger$} & 72.87 & 0.9200 & 13.52 & 3.344 & 0.1200 & 0.253 & 0.00439 & \underline{14.2} & \textbf{48} \\
\midrule
\sysname{} Stage 2 & 50 & \underline{75.15} & \underline{0.9476} & \underline{16.69} & \underline{1.308} & \underline{0.0381} & \textbf{0.157} & \underline{0.00199} & 4.4 & \underline{54} \\
\sysname{} Stage 3 & 4 & 74.52 & \textbf{0.9610} & \textbf{18.31} & \textbf{1.075} & \textbf{0.0318} & \underline{0.159} & \textbf{0.00196} & \textbf{20.6} & \underline{54} \\
\bottomrule

\end{tabular}
}
%\vspace{-4mm}
\par\smallskip
{\footnotesize\raggedright
\textsuperscript{$\dagger$}We follow the official SANA-WM inference
implementation, using four denoising steps followed by
three refinement steps.\par}
\label{tab:main}
\end{table*}

\subsection{Revisit Fidelity and Camera Control}
\label{sec:res_camera}

\textbf{Revisits beyond the Recent Context.}
Our \sysname significantly improves both semantic return consistency and pixel-level reference fidelity (Table~\ref{tab:main}). Compared with SANA-WM, the strongest baseline on revisit fidelity, our Stage~3 model roughly halves the DINOv2 feature distance (reducing error from $0.0800$ to $0.0390$) and raises PSNR by $4.79$\,dB. Both gains are supported by paired sign tests (Appendix~\ref{app:paired_results}). These two metrics provide complementary evidence: outbound-return similarity measures the retention of generated content, whereas GT-referenced PSNR measures agreement with the reference scene. Their joint improvement demonstrates superior faithful scene recovery ability of our method beyond the active context. Additional comprehensive evaluations across DINOv2, PSNR, SSIM, and LPIPS (Appendix~\ref{app:metric_robustness}) show that both our Stage~2 teacher and Stage~3 student consistently outperform all baselines on the revisit measures.

\textbf{Trajectory Following.}
Faithful scene revisits inherently necessitate precisely reaching the requested camera viewpoint. Relative to SANA-WM, our \sysname Stage~3 reduces absolute RotErr from $3.344^\circ$ to $1.075^\circ$, alongside lower rotational and translational RPE (Table~\ref{tab:main}). Notably, several baselines exhibit small consecutive-frame errors with markedly higher absolute errors, a pattern reflecting accumulated trajectory drift. The lower errors of \sysname on both local and global scales indicate more coherent local motion and overall camera following. Together with the revisit results, these findings show improved scene recovery while executing the prescribed camera excursion.

\textbf{Qualitative Revisits.}
Figure~\ref{fig:revisit-memory} compares our \sysname{} with representative baselines on out-and-back camera trajectories. In the indoor scene featuring complex geometries (\textit{e.g.}, staircases, windows, and reflections), and the outdoor environment containing fine structures, comparing the initial and return views reveals that existing baselines often suffer from changes in scene layout, object appearance, and viewpoint upon revisiting. In contrast, our method accurately retains object identities and structural alignment. Additional qualitative results along 30-second real-world camera trajectories are provided in Appendix~\ref{app:qualitative_revisits}.

\begingroup
\setlength{\intextsep}{0pt}
\setlength{\columnsep}{12pt}
\begin{wraptable}{r}{0.5\textwidth}
\vspace{0pt}
\centering
\footnotesize
\setlength{\tabcolsep}{3pt}
\caption{\textbf{Revisit Consistency under Increasing Camera Sweep.} T1--T3 are equal-sized terciles ($40$ clips each) of yaw sweep normalized by horizontal field of view. Higher tiers indicate larger sweeps. Bold and underline mark the best and second-best generated similarities in each tier, respectively.}
\begin{tabular}{l ccc c}
\toprule
Method & T1 & T2 & T3 & T3$-$T1 \\
\midrule
GT (VAE-decoded) & 0.9981 & 0.9983 & 0.9981 & $0.000$ \\
\midrule
Matrix-Game 3.0 & 0.8743 & 0.8196 & 0.7737 & $-0.101$ \\
HY-World 1.5 & 0.8612 & 0.8581 & 0.8497 & $-0.012$ \\
DreamX-AR & 0.9074 & 0.8941 & 0.8289 & $-0.079$ \\
LingBot-Fast & 0.8532 & 0.8701 & 0.8506 & $-0.003$ \\
SANA-WM & 0.9201 & {0.9311} & 0.9090 & $-0.011$ \\
\midrule
\sysname{} Stage 2 & \underline{0.9434} & \underline{0.9501} & \underline{0.9494} & $+0.006$ \\
\sysname{} Stage 3 & \textbf{0.9582} & \textbf{0.9644} & \textbf{0.9603} & $+0.002$ \\
\bottomrule
\end{tabular}
\captionsetup{font=footnotesize,skip=4pt}
\label{tab:difficulty}
\end{wraptable}

\textbf{Larger Camera Sweeps.}
\label{sec:difficulty}
We partition the benchmark clips post-hoc into three equal-sized tiers based on total yaw sweep normalized by horizontal field of view, serving as a proxy for the volume of spatial content that leaves the active field of view. This stratification relies only on the prescribed camera trajectories.
As reported in Table~\ref{tab:difficulty}, our Stage~3 student model consistently maintains high revisit semantic similarity across all sweep tiers. In contrast, Matrix-Game 3.0 and DreamX-AR significantly degrade under larger sweeps, while other baselines remain lower in absolute similarity despite flatter trends. Notably, our student model achieves advantages over SANA-WM across all tiers, with the most pronounced performance margin observed in the highest-sweep tier (T3). The concurrently lower camera-following error of our student model in this challenging tier further supports its robust and accurate revisit under substantial viewpoint changes (Appendix~\ref{app:difficulty_details}).

\par
\endgroup

\subsection{Step Distillation and Efficiency}
\label{sec:distillation_results}

Stage~3 distillation (\S\ref{sec:stage3}) drastically reduces sampling from $50$ to $4$ steps while retaining strong revisit performance. The student model increases revisit PSNR from $16.69$ to $18.31$\,dB and improves revisit LPIPS, with both gains supported by paired statistical tests (Appendix~\ref{app:metric_robustness}). Its mean DINOv2 revisit distance is also lower, achieving $72/120$ paired wins ($p=0.0353$) against the teacher. Paired tests detect no statistically significant discrepancy between the teacher and student models across all four camera metrics (Appendix~\ref{app:distillation_details}). The primary trade-off is in visual quality and local consistency: VBench decreases from $75.15$ to $74.52$, and Local Memory from $0.9442$ to $0.9258$. The 4-step student therefore combines enhanced pixel-level revisit fidelity with modest compromises in these localized perceptual scores.

To enable real-time interactive generation, we combine memory-read caching, compilation, efficient attention kernels, and optimized VAE decoding to reduce the end-to-end latency and peak memory of the model inference (Appendix~\ref{app:inference}). With online TTT memory updates active, the 4-step student sustains an end-to-end throughput of $20.6$\,FPS at $1280\times704$ on a single NVIDIA B200 GPU (Table~\ref{tab:main}, Figure~\ref{fig:teaser}), exceeding the $16$\,FPS real-time rate. The fixed-size parametric memory and bounded recent context keep memory storage and per-step costs independent of rollout duration (\S\ref{sec:stage2}).

\section{Conclusion}
\label{sec:conclusion}
We presented \sysname, a camera-controlled world model that consolidates historical observations into a parametric memory by test-time training (TTT).
To address spatial redundancy along smooth camera trajectories, we introduced an online recursive least-squares update rule that selectively writes novel visual contexts while preventing stored scene memory from degradation.
We instantiated \sysname through progressive three-stage training that transitions a pretrained video diffusion model into an autoregressive world model with TTT memory, followed by dual-teacher distillation for few-step generation.
Evaluations show that \sysname reliably preserves persistent world state, achieving superior revisit consistency and sustaining $20.6$ FPS at $1280\times704$ on one NVIDIA B200 GPU.

% \clearpage
% \newpage
\bibliographystyle{assets/plainnat}
\bibliography{paper}

\clearpage
\newpage
\beginappendix
\section{Additional Qualitative Revisit Results}
\label{app:qualitative_revisits}

\begin{figure}[htbp]
    \centering
    \includegraphics[width=\linewidth]{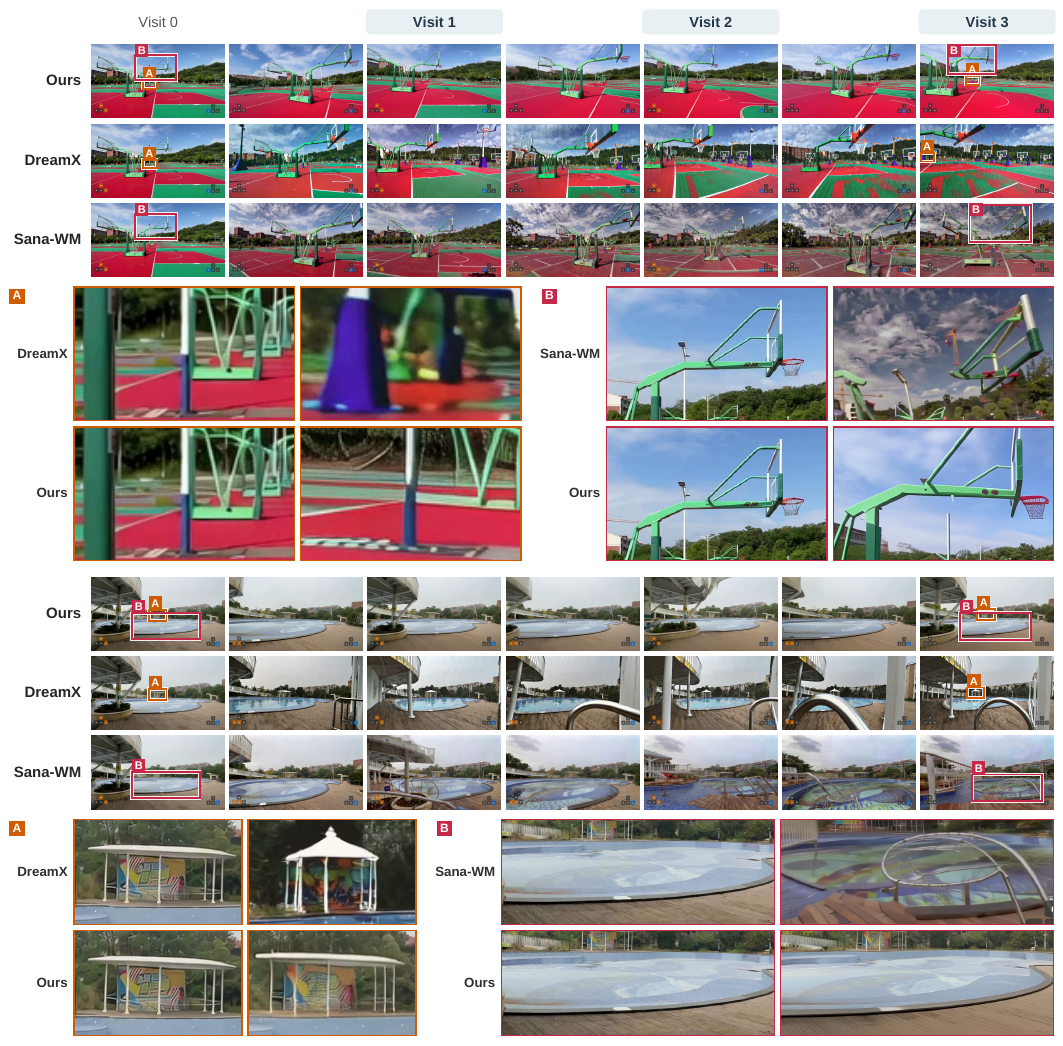}
    \vspace{-4mm}
    \caption{\textbf{Scene Consistency across Approximate Revisits.} We identify three approximate revisits within a 30-second real camera motion sequence. Unlike the prescribed out-and-back trajectory used for quantitative evaluation, these revisits observe the same scene regions from similar, rather than identical, camera poses. Visit 0 denotes the initial observation, and Visits 1, 2, and 3 mark the selected revisits in temporal order. Colored boxes link full views to enlarged crops comparing local details in the baselines and \sysname. All enlarged crops compare Visits 0 and 3.
    }
    \label{fig:revisit-memory-30s}
\end{figure}

Figure~\ref{fig:revisit-memory-30s} provides a qualitative comparison along a 30-second real camera motion sequence, during which \sysname continually updates its scene memory through TTT. We select three approximate revisits and enlarge local details to inspect scene retention. The 10-second quantitative benchmark uses a prescribed out-and-back trajectory to pair views at the same requested pose. Here, the selected views revisit the same scene regions without exact pose matching. At the basketball court scene, \sysname{} retains the slim blue lower section of the lamp post, whereas DreamX-AR produces a broad purple pedestal (crop A). SANA-WM disconnects a backboard from its support (crop B). At the pool scene, DreamX-AR changes the pavilion's flat roof to a pointed roof, and SANA-WM alters the pool-surface pattern. In contrast, our \sysname more consistently retains these structures and appearance cues across visits.

\section{Training Data}
\label{sec:data}

Learning a camera-controllable video world model that maintains persistent scene memory imposes four complementary requirements on the training corpus: visual richness and diversity, precise camera pose annotations to ground the control signal, sequence lengths commensurate with the target generation horizon, and viewpoint revisits, whereby a previously observed region must be faithfully recalled even after substantial intervening content has elapsed beyond the model's context. To satisfy these desiderata, we curate a base collection of 601K clips that provides broad coverage of appearance and camera motion and an augmented set of 307K clips designed for viewpoint recurrence.

\subsection{Corpus Composition}
\label{sec:composition}

Following recently released work in video world models \citep{sanawm2026, dreamxworld2026}, our training corpus draws on both real-world footage and synthetic renderings to jointly cover natural appearance, diverse camera dynamics, and precise geometric annotation. Every clip is preprocessed into a unified data structure comprising video latents, per-frame camera intrinsics and extrinsics, and a free-form language scene caption, then resampled to $1280\times704$ resolution at $16$\,FPS.

\textbf{Real-World Sources.}
Captured video provides the naturalistic appearance and complex illumination. Our real-world subset spans indoor and outdoor environments, encompassing casual orbiting and sweeping motions around objects or scenes, egocentric urban traversal through city streets, and smooth architectural walkthroughs of interior spaces. Together, these sources furnish a broad distribution over both scene category and camera trajectory, with an emphasis on forward-moving, first-person viewpoints in open environments.

\textbf{Synthetic Sources.}
Game-engine renderings complement real capture along two axes. First, they supply ground-truth camera pose free of estimation error, serving as a noise-free supervision signal. Second, they introduce trajectory patterns such as sustained aerial flight and systematic ground-level exploration that are difficult to acquire in the physical world, thereby extending the motion distribution covered by the corpus.

\subsection{Camera Annotation}
\label{sec:annotation}
To obtain coherent camera parameterization, we apply VGGT-$\Omega$~\citep{vggtomega2026} to all training clips, recovering per-frame intrinsics and extrinsics within a unified scale and anchoring each trajectory to the initial frame as the world origin. The resulting camera trajectories serve as the conditioning signal for controllable video generation and further induce a per-clip co-visibility matrix that encodes pairwise surface overlap between frames, quantifying the degree of viewpoint revisit throughout each sequence.

\subsection{Data Filtering}
\label{sec:filtering}
We filter each clip through a two-stage pipeline: objective criteria drawn from the imagery and the recovered geometry, and a semantic assessment produced by vision-language models.

\textbf{Objective Scoring.}
The objective criteria span both visual content and camera geometry, following the conventions established for camera-controllable corpora~\citep{sanawm2026,lingbotworld2026, wonder2026, alayaworld2026}. Perceptual quality is assessed with DOVER~\citep{dover2023}, which produces complementary technical and aesthetic scores: the technical score captures low-level distortions such as blur and artefacts, while the aesthetic score reflects content and composition. To characterize scene dynamics, we compute optical flow magnitude via UniMatch~\citep{unimatch2023} between frames and impose bilateral bounds. Clips whose flow falls below a lower threshold convey negligible information per unit duration, whereas those exceeding an upper threshold suffer from motion blur and inter-frame disocclusion. Mean luminance is similarly constrained to a permissible band, eliminating under-exposed footage in which scene structure is irrecoverable and over-exposed footage in which highlight detail is clipped. On the geometric side, we retain only clips whose estimated field of view and focal variation remain within the regime where the Plücker ray representation of \S\ref{sec:stage1} provides a faithful camera embedding.

\textbf{Semantic Profiling.}
Beyond photometric and geometric criteria, we employ a vision-language model to perform semantic assessment. Video frames are sampled uniformly across each clip and presented to the model, which annotates the scene along several aspects: the count of dynamic entities (pedestrians, vehicles, animals), the presence of synthetic overlays (UI chrome, loading screens, watermarks), perceptual blur, and viewpoint characterization. Responses conform to a structured schema in which factual observations and a concise rationale precede the binary usability verdict, enforcing a chain-of-thought discipline that grounds the decision in explicit evidence. Final admission into the corpus is gated jointly on the entity count and the usability label.

\subsection{Text Captioning}
\label{sec:captioning}
Captions serve as the textual conditioning signal and are generated by a vision-language model prompted to produce scene-static descriptions~\citep{lingbotworld2026,sanawm2026} in a single paragraph. A scene-static caption characterizes the environment in terms of its constituent objects, spatial arrangement, materials, colour palette, and illumination, together with dynamics intrinsic to the scene itself, such as flowing water or drifting fog. Crucially, the caption abstracts away all camera-related language: viewpoint references, cinematographic terminology, and motion verbs such as \emph{pan}, \emph{dolly}, or \emph{fly over} are systematically excluded. The guiding principle is invariance: a given scene should elicit semantically equivalent captions regardless of the trajectory along which it is observed, ensuring that camera motion is communicated exclusively through the pose conditioning branch and that the two modalities of control remain disentangled.

\subsection{Revisit Construction}
\label{sec:revisits}

Revisits, denoting returns to previously observed regions, arise sparingly in natural video, since a camera traversing a scene seldom retraces its path. As a result, the co-visibility statistic remains low across most of the base set. This scarcity hinders learning an effective memory mechanism, which relies on repeated observations of the same content.
To expose the model to such recurrence in real-world data, we construct revisits through a clip replay procedure.
Specifically, we loop and re-index the camera trajectory of a video sequence so that the temporal path periodically returns to previously observed viewpoints at distinct temporal intervals within the same training sequence.
We confine the replay to static scenes and exclude dynamic entities that change between visits to avoid abnormal movements. For synthetic data, we further design camera trajectories as closed loops through the rendered environments, so that the viewpoint deliberately returns to previously visited locations. Taken together, these procedures yield 307K clips with recurring viewpoints.

\section{Training Details}
\label{app:training}
The three stages of our method are implemented as five consecutive training runs: bidirectional camera-conditioned supervised fine-tuning (SFT), block-causal adaptation, memory-augmented training, consistency distillation (CD), and distribution matching distillation (DMD).
Specifically, the first run constitutes Stage~1, the subsequent two runs comprise Stage~2, and the final two runs complete Stage~3.
Table~\ref{tab:training_schedule} outlines the training configurations across all runs.

\subsection{Shared Configuration}
\label{app:training_shared}

We initialize from Wan2.2-TI2V-5B, whose diffusion transformer (DiT) consists of $30$ layers, a hidden dimension of $3072$, and $24$ attention heads with a head dimension of $128$.
All training runs use $1280\times704$ videos at $16$\,FPS and BF16 mixed precision.
The frozen WanVideoVAE compresses raw input videos into $48$-channel latents with $16\times$ spatial and $4\times$ temporal downsampling factors.
Specifically, a video clip of $T$ frames maps to $F = 1 + (T-1)/4$ latent frames over a $44\times80$ spatial grid, with the DiT processing patches of size $(1,2,2)$.
Accordingly, $21$ and $63$ latent frames correspond to $81$ and $249$ raw video frames, respectively.
Unless otherwise specified, all temporal metrics below, including context windows, sinks, block sizes, and rollout lengths, are quantified in terms of latent frames.

The initial video frame conditions generation via its clean latent at diffusion timestep zero, which is directly concatenated along the temporal sequence.
The standard denoising loss excludes this clean initial frame.
Textual conditioning is injected via cross-attention over cached T5 embeddings, with both the text encoder and VAE remaining frozen throughout training.
Camera control is parameterized using Pl\"ucker raymaps injected through camera-dependent adaptive layer normalization (CamAdaLN).

Optimization is performed using AdamW across all training runs, adhering to the stage-specific schedules listed in Table~\ref{tab:training_schedule}.
With a gradient accumulation factor of one, the effective global batch size equals the per-GPU batch size multiplied by the number of active data-parallel workers.

\begin{table}[t]
\centering
\small
\setlength{\tabcolsep}{3pt}
\caption{\textbf{Training Configuration.} ``AR'' denotes block-causal autoregressive adaptation, and ``Memory'' represents the subsequent memory-augmented fine-tuning stage. Learning rates correspond to peak values, and gradient clipping thresholds are specified as gradient norms. DMD iteration counts represent critic updates, with generator updates executed every five iterations.}
\label{tab:training_schedule}
\begin{tabular}{lccccc}
\toprule
 & Stage 1 & \multicolumn{2}{c}{Stage 2} & \multicolumn{2}{c}{Stage 3} \\
\cmidrule(lr){3-4}\cmidrule(lr){5-6}
 & SFT & AR & Memory & CD & DMD \\
\midrule
Training iterations & $6000$ & $10{,}000$ & $15{,}000$ & $2500$ & $800$ \\
Latent frames & $21$ & $21$ & $63$ & $63$ & $63$ \\
GPUs & $512$ & $256$ & $128$ & $128$ & $128$ \\
Batch per GPU & $2$ & $2$ & $1$ & $1$ & $1$ \\
Global batch & $1024$ & $512$ & $128$ & $128$ & $128$ \\
Backbone/generator LR & $5\!\times\!10^{-5}$ & $5\!\times\!10^{-5}$ & $5\!\times\!10^{-5}$ & $2\!\times\!10^{-6}$ & $2\!\times\!10^{-6}$ \\
Memory LR multiplier & -- & -- & $15$ & $1$ & -- \\
Critic LR & -- & -- & -- & -- & $4\!\times\!10^{-7}$ \\
Warm-up iterations & $500$ & $500$ & $500$ & $200$ & $50$ \\
LR floor / peak & $0.01$ & $0.1$ & $0.1$ & $0.1$ & $0.1$ \\
AdamW $\beta_1$ & $0.9$ & $0.9$ & $0.9$ & $0$ & $0$ \\
AdamW $\beta_2$ & $0.999$ & $0.999$ & $0.999$ & $0.999$ & $0.999$ \\
Weight decay & $0.01$ & $0.01$ & $0.01$ & $0$ & $0$ \\
Gradient clipping & $0.5$ & $0.5$ & $0.5$ & $0.5$ & $1.0$ \\
\bottomrule
\end{tabular}
\end{table}

\subsection{Stage 1: Bidirectional Camera-Conditioned SFT}
\label{app:training_sft}

In Stage~1, we train rank-stabilized LoRA adapters of rank $r=256$ and scaling factor $\alpha=16$ on all attention projections (query, key, value, output) and feed-forward networks, alongside rank-$512$ CamAdaLN modules.
All remaining pretrained backbone weights are held frozen.
Training optimizes a camera-conditioned denoising loss with full bidirectional attention across sequences of $21$ latent frames.
To enable classifier-free guidance, camera conditioning is randomly dropped with a probability of $0.05$, with the dropout decision synchronized across the batch.
We employ distributed data parallelism with AdamW.
The learning rate linearly warms up for $500$ iterations to a peak of $5\times10^{-5}$, followed by cosine decay toward a floor of $5\times10^{-7}$ over the total horizon.

\subsection{Stage 2: Autoregressive Adaptation and Memory Training}
\label{app:training_ar_memory}

\textbf{Block-Causal Adaptation.}
Starting from the Stage~1 checkpoint, we continue training the same LoRA adapters and camera modules for $10{,}000$ iterations using masked teacher forcing.
Attention is converted to block-causal, with three latent frames per block, and each target block attends to itself and preceding context within a $21$-latent window.
We retain distributed data parallelism, AdamW, and the peak learning rate of $5\times10^{-5}$.
After $500$ warm-up iterations, cosine decay reaches $5\times10^{-6}$ at iteration $10{,}000$.

\textbf{Memory-Augmented Fine-Tuning.}
We merge the iteration-$10{,}000$ LoRA weights into the backbone, instantiate the TTT memory modules, and jointly fine-tune the entire DiT backbone, camera modules, and TTT memory parameters for $15{,}000$ iterations.
The memory parameter group uses a learning-rate multiplier of $15$, giving a peak learning rate of $7.5\times10^{-4}$, while the backbone peak remains $5\times10^{-5}$.
The backbone schedule uses $500$ warm-up iterations followed by global-step cosine decay with a floor of $5\times10^{-6}$.
We employ FSDP2 full sharding for the DiT backbone and replicate the memory modules across workers.

Each training episode contains $63$ latent frames, with three latent frames per autoregressive block, a $21$-latent sliding attention window, and a persistent three-latent sink block.
Memory writes use the recursive least-squares readout update (\texttt{rls\_w2}) with regularization $\lambda=10^{-3}$, performing $14$ online memory consolidation updates per episode.
Keys and values from clean context blocks are cached and detached from the compute graph.
Revisit clips are constructed by decoding an $81$-frame source segment, alternating forward and reverse traversal to form $249$ frames, and re-encoding to $63$ latents, with camera poses following the same temporal indexing.

\subsection{Stage 3: Few-Step Distillation}
\label{app:training_distill}

\textbf{Consistency Distillation.}
We initialize the student model weights from the Stage~2 memory teacher, and jointly train DiT backbone, camera modules, and TTT memory parameters for $2500$ iterations on the same training corpus in this stage.
Consistency Distillation (CD) retains the configuration from the previous stage, including $63$-latent episodes, three-latent blocks, $21$-latent window, three-latent sink, and the RLS memory update rules.
The generation schedule employs a discretization grid of $N=48$, with a 4-step sampling schedule evaluated at step indices $[0,12,24,36]$.
The teacher guidance scale is fixed to $1.0$.
All trainable parameter groups, including the memory modules, adopt a peak learning rate of $2\times10^{-6}$, with $200$ warm-up iterations and cosine decay to $10\%$ of the peak value.
Gradient norms are clipped at $0.5$, with a gradient-spike threshold set to $3.0$.
We use FSDP2, seed $0$, and an exponential moving average (EMA) with a decay factor of $0.99$, starting at iteration $200$ and updated every iteration.
The iteration-$2500$ EMA checkpoint serves as the initializer for the DMD generator.

\textbf{Dual-Teacher Distribution Matching Distillation.}
DMD initializes the generator weights from the trained CD model while resetting the optimizer state. It incorporates both the frozen Stage~1 bidirectional teacher and the frozen Stage~2 memory-augmented teacher to provide dual-teacher guidance.
Each teacher is paired with a trainable critic network possessing an identical architecture, initialized via weight copies.
Both memory-augmented scoring networks construct their fast-weight memory states directly from the student's own generated rollouts.

The DMD training phase spans $800$ iterations.
Iterations $1$--$399$ use only the bidirectional score term, while iterations $400$--$800$ use guidance from both the bidirectional and memory-augmented teachers.
At iteration $400$, the generator learning rate undergoes an additional $50$-step boundary warm-up.
Critics are updated every iteration, whereas the generator updates every fifth iteration, yielding $160$ generator optimization steps across the run.
The peak learning rates are $2\times10^{-6}$ for the generator and $4\times10^{-7}$ for the critics, with $50$ initial warm-up iterations and cosine decay to $10\%$ of each peak value.
AdamW hyperparameters are set to $(\beta_1,\beta_2)=(0,0.999)$ with zero weight decay and a gradient clipping threshold of $1.0$.
We use FSDP2, seed $0$, and generator EMA with decay $0.99$ starting from iteration $200$.

For each training iteration, the rollout length is sampled uniformly from $\{21,24,\ldots,63\}$ latent frames and synchronized across parallel workers.
The DMD loss is applied to the final contiguous $21$ latent frames, using the same 4-step denoising schedule as CD.
The final model is taken from the iteration-$800$ EMA checkpoint after the dual-teacher distillation phase.

\section{Long-Horizon Revisit: Protocol and Additional Results}
\label{app:replay}

\subsection{Evaluation Set and Shared Preprocessing}
\label{app:replay_protocol}
The revisit evaluation benchmark consists of $120$ video clips curated from an unseen, held-out test split of our dataset.
For rigorous and unbiased evaluation of memory retention, clip selection is governed by a filtering pipeline based on trajectory geometry.
Specifically, we enforce a geometric constraint requiring a normalized angular yaw sweep (total sweep relative to horizontal field of view) of at least $0.30$.
The resulting benchmark exhibits a median angular sweep of $38^\circ$ and a median sweep/FOV ratio of $0.46$.
To guarantee fair representation, clips are stratified across source domains according to their natural proportions in the dataset pool, mitigating motion distribution bias and avoiding overrepresentation of specific camera dynamics.

All evaluated methods are conditioned on identical per-clip random seeds, text prompts, and prescribed 6-DoF camera trajectories.
Every sequence is generated and evaluated for exactly $160$ frames at $16$\,FPS and $1280\times704$ resolution. We enforce strict frame-count truncation (rather than time-based truncation) to ensure uniform evaluation sequence lengths.
Videos are encoded using standard H.264 (CRF $12$, yuv420p).
The ground-truth (GT) reference is decoded from $41$ cached VAE latents ($161$ frames) and truncated to $160$ frames.
Note that GT passes through one video-encoding stage, whereas generated outputs are re-encoded from their native videos and thus pass through two. This minor encoding asymmetry can slightly favor GT.

\subsection{Metric Definitions}
\label{app:replay_metrics}
\textbf{Revisit Similarity.}
For a generated video clip $x^m$, let $f(I)$ denote the $L_2$-normalized mean of non-CLS patch tokens extracted from DINOv2 ViT-B/14.
Input images are resized to a short side of $256$, center-cropped to $224\times224$, and normalized using standard ImageNet statistics.
Given the zero-indexed frame pairs $\mathcal P = \{(8,152), (16,144), (24,136),$ $ (32,128)\}$, we define:
\begin{equation}
d(I,I')=1-f(I)^\top f(I'),\qquad
\bar d_m=\frac{1}{4}\sum_{(a,b)\in\mathcal P}d(x^m_a,x^m_b),\qquad
S_m=1-\bar d_m.
\label{eq:replay_dino}
\end{equation}
We report the aggregate DINOv2 similarity averaged across all $120$ benchmark clips.
The temporal separations between these paired frames correspond to $36$, $32$, $28$, and $24$ latent frames, respectively, all strictly exceeding the model's $21$-latent-frame active context window.
This explicitly tests the model's ability to recall return views beyond its transient attention context.
Our protocol adapts WorldMark's revisit evaluation framework~\citep{worldmark2026}, but employs fixed-index frame pairs rather than optical-flow-matched pairs.
Fixed-time pairing evaluates the commanded camera trajectory and intentionally retains camera-following errors.
To complement self-consistency, we evaluate each generated outbound and return view against its corresponding VAE-decoded ground-truth reference $g$, averaging over the same four pairs:
\begin{equation}
d_{\rm first} = \frac{1}{4}\sum_{(a,b)\in\mathcal P} d(x^m_a, g_a),\qquad
d_{\rm return} = \frac{1}{4}\sum_{(a,b)\in\mathcal P} d(x^m_b, g_b).
\end{equation}
We average these per-clip scores over the $120$ clips and report $\Delta=d_{\rm return}-d_{\rm first}$, the mean increase in reference distance between the initial and return visits.
These reference-based comparisons and camera trajectory errors distinguish faithful memory retention from self-consistent but incorrect scene content.
Center cropping additionally restricts metric sensitivity to peripheral image boundary artifacts.

\textbf{Pixel Fidelity.}
For each outbound-return pair $(a,b)\in\mathcal{P}$, we compute revisit PSNR and SSIM between the generated return view $x^m_b$ and its VAE-decoded reference $g_b$ at the full $1280\times704$ resolution.
Scores are averaged across the four pairs per clip and subsequently aggregated across the benchmark.
For RGB values normalized to $[0,1]$, $\mathrm{PSNR} = 10\log_{10}(1/\mathrm{MSE})$, with MSE values below $10^{-12}$ capped at $99$\,dB.
SSIM is calculated via \texttt{skimage.metrics.structural\_similarity} with \texttt{channel\_axis=2}, \texttt{data\_range=1.0}, and default $7\times7$ Gaussian windowing.
First-visit PSNR is computed analogously between $x^m_a$ and $g_a$ over the same four pairs.
Our 4-step student model achieves a first-visit PSNR of $19.99$\,dB and a return-visit PSNR of $18.31$\,dB (a modest degradation of $1.68$\,dB).
We additionally report Self-SSIM averaged across the generated outbound and return views $(x^m_a, x^m_b)$.
Crucially, pixel-level metrics penalize camera control drift alongside appearance degradation, serving as a holistic measure of spatial and structural alignment.

\textbf{Camera Following.}
Camera poses are extracted from all generated videos and GT sequences using the VGGT-$\Omega$~\citep{vggtomega2026}, enforcing $160$ reconstructed poses per video.
Recovered and prescribed camera-to-world poses are expressed relative to frame $0$.
We align translations with a scalar $s^*=\sum_t\mathbf t_t^\top\widetilde{\mathbf t}_t/\sum_t\|\widetilde{\mathbf t}_t\|_2^2$, without fitting global rotation or translation offsets.
This scale-only convention preserves the shared initial-frame coordinate system and avoids unstable rotational alignment for nearly planar trajectories.
Let $P_t=[R_t,\mathbf t_t;\mathbf 0^\top,1]$ and $\widetilde P_t$ denote the reference and scale-aligned recovered poses.
With $\theta(R)=\frac{180}{\pi}\arccos(\operatorname{clip}((\operatorname{tr}R-1)/2,-1,1))$, absolute trajectory errors are defined as:
\begin{equation}
\mathrm{RotErr}=\frac{1}{T}\sum_t\theta(R_t^\top\widetilde R_t),\qquad
\mathrm{CamMC}=\frac{1}{T}\sum_t\|P_t-\widetilde P_t\|_F,\quad T=160.
\end{equation}
For consecutive frames, set $A_t=P_t^{-1}P_{t+1}$, $\widetilde A_t=\widetilde P_t^{-1}\widetilde P_{t+1}$, and $E_t=A_t^{-1}\widetilde A_t$.
Relative Pose Errors (RPE) for rotation and translation are computed as the average of $\theta(E_t[1{:}3,1{:}3])$ and $\|E_t[1{:}3,4]\|_2$ across $T-1$ adjacent frame pairs, respectively.
Rotation metrics apply no alignment, whereas translation metrics employ the same scalar factor $s^*$.
Following temporal averaging per clip, dataset-macro averages are reported across the source domains.
The non-zero error of GT in Table~\ref{tab:main} quantifies the intrinsic baseline noise of the pose estimator.
Absolute errors reveal accumulated pose drift, whereas RPE measures consecutive-frame motion.

\textbf{Visual Quality and Aggregation.}
We use the unmodified VBench 0.1.5 evaluator~\citep{vbench2024} in \texttt{custom\_input} mode with its default temporal sampling and image preprocessing.
Table~\ref{tab:replay_vbench} reports all seven dimensions on a $0$--$100$ scale: imaging quality, aesthetic quality, subject consistency, background consistency, motion smoothness, temporal flickering, and overall consistency, the last evaluated against the source-video text prompts.
Each dimension is the mean of per-video scores over all benchmark clips, and the unweighted average of the seven dimensions is the VBench score reported in Table~\ref{tab:main}.
Note that VBench downsamples images to $512$ on the longest side for imaging quality and evaluates aesthetic quality on $224\times224$ center crops, reflecting perceptual appeal rather than native-resolution boundary fidelity.

\textbf{Local Temporal Consistency.}
Inspired by WorldMark, we additionally report Local Memory score $\mathrm{LM} = (1-r_{\rm cut})(1-r_{\rm mut})$.
Hard scene cuts are detected using TransNetV2 at threshold $0.5$, while visual mutations are defined as adjacent sampled frames with a DINOv2 feature distance exceeding $0.10$.
Frames are sampled at fixed $0.5$\,s intervals (excluding pairs within $0.5$\,s of a detected cut).
LM evaluates localized temporal smoothness and short-term frame transitions.

\subsection{Additional Results and Trade-offs}
\label{app:replay_results}

% \begin{table}[t]
% \centering
% \small
% \setlength{\tabcolsep}{5pt}
% \caption{\textbf{Visual Quality on the Revisit Benchmark.} All dimensions are scored on a $0$--$100$ scale (higher is better). VB-4 represents the unweighted mean of Imaging, Aesthetic, Motion, and Flicker scores. Bold indicates the top performance.}
% \begin{tabular}{l rrrrr}
% \toprule
% Method & Imaging & Aesthetic & Motion & Flicker & VB-4 \\
% \midrule
% GT (VAE-decoded) & 74.00 & 53.84 & 98.27 & 94.37 & 80.12 \\
% \midrule
% Matrix-Game 3.0 & 74.03 & 50.66 & 98.42 & 94.59 & 79.42 \\
% HY-World 1.5 & \textbf{76.29} & 52.44 & 97.79 & 93.82 & \textbf{80.09} \\
% DreamX-AR & 73.59 & 52.49 & 97.67 & 94.48 & 79.56 \\
% LingBot-Fast & 74.13 & \textbf{52.84} & 97.08 & 94.62 & 79.67 \\
% SANA-WM & 65.21 & 47.04 & \textbf{98.86} & \textbf{95.40} & 76.63 \\
% \midrule
% \sysname Stage 2 & 73.17 & 51.96 & 98.35 & 94.70 & 79.55 \\
% \sysname Stage 3 & 71.46 & 50.14 & 98.47 & 95.28 & 78.83 \\
% \bottomrule
% \end{tabular}
% \label{tab:replay_vbench}
% \end{table}

\begin{table}[t]
\centering
\footnotesize
\setlength{\tabcolsep}{4.5pt}
\caption{\textbf{Visual Quality on the Revisit Benchmark.} All seven VBench dimensions are scored on a $0$--$100$ scale (higher is better). ``Avg.'' is their unweighted mean and equals the VBench column of Table~\ref{tab:main}. Bold indicates the best generated result and underline indicates the second best.}
\begin{tabular}{l ccccccc c}
\toprule
Method & Imaging & Aesthetic & Subject & Background & Motion & Flicker & Overall & Avg. \\
\midrule
GT (VAE-decoded) & 74.00 & 53.84 & 91.27 & 93.44 & 98.27 & 94.37 & 24.65 & 75.69 \\
\midrule
Matrix-Game 3.0 & 74.03 & 50.66 & 89.53 & 91.81 & 98.42 & 94.59 & 24.00 & 74.72 \\
HY-World 1.5 & \textbf{76.29} & 52.44 & \textbf{90.72} & 92.34 & 97.79 & 93.82 & 23.87 & \textbf{75.32} \\
DreamX-AR & 73.59 & \underline{52.49} & 88.53 & 91.99 & 97.67 & 94.48 & \underline{25.05} & 74.83 \\
LingBot-Fast & \underline{74.13} & \textbf{52.84} & 88.97 & 91.97 & 97.08 & 94.62 & \textbf{25.22} & 74.98 \\
SANA-WM & 65.21 & 47.04 & 88.40 & 91.46 & \textbf{98.86} & \textbf{95.40} & 23.75 & 72.87 \\
\midrule
\sysname Stage 2 & 73.17 & 51.96 & \underline{90.56} & \textbf{92.99} & 98.35 & 94.70 & 24.29 & \underline{75.15} \\
\sysname Stage 3 & 71.46 & 50.14 & 90.04 & \underline{92.56} & \underline{98.47} & \underline{95.28} & 23.67 & 74.52 \\
\bottomrule
\end{tabular}
\label{tab:replay_vbench}
\end{table}

As detailed in Tables~\ref{tab:main} and~\ref{tab:replay_vbench}, the student model's revisit fidelity gains coexist with slightly lower VBench visual quality scores compared to its teacher.
Notably, the decoded GT reference achieves an imaging quality score of $74.00$, lower than several generated baselines, highlighting that VBench imaging quality measures stylized sharpness rather than absolute distance to real-world ground truth.

\begin{table}[t]
\centering
\footnotesize
\setlength{\tabcolsep}{4.5pt}
\caption{\textbf{Complementary Revisit Consistency Measures.} $\bar d$ denotes the DINOv2 self-distance across four outbound-return pairs. $d_{\rm first}$ and $d_{\rm return}$ evaluate feature distance against GT for outbound and return frames, respectively ($\Delta = d_{\rm return}-d_{\rm first}$). Self-SSIM compares generated return views to outbound views, and GT-SSIM compares return views to GT references. Bold marks the best generated result and underline denotes the second best.}
\begin{tabular}{l ccccccc}
\toprule
Method & $\bar d\downarrow$ & $d_{\rm first}\downarrow$ & $d_{\rm return}\downarrow$ & $\Delta\downarrow$ & Self-SSIM$\uparrow$ & GT-SSIM$\uparrow$ & LM$\uparrow$ \\
\midrule
GT (VAE-decoded) & 0.0018 & -- & -- & -- & 0.9470 & -- & 0.9756 \\
\midrule
Matrix-Game 3.0 & 0.1775 & 0.0411 & 0.2185 & 0.1774 & 0.3759 & 0.3748 & 0.8875 \\
HY-World 1.5 & 0.1437 & \underline{0.0274} & 0.1867 & 0.1593 & 0.3576 & 0.3526 & \textbf{0.9874} \\
DreamX-AR & 0.1232 & 0.0327 & 0.1610 & 0.1282 & 0.3805 & 0.3781 & \underline{0.9838} \\
LingBot-Fast & 0.1420 & 0.0279 & 0.1832 & 0.1553 & 0.3431 & 0.3409 & 0.9703 \\
SANA-WM & 0.0800 & 0.0587 & 0.1218 & 0.0631 & 0.4704 & 0.4360 & 0.9006 \\
\midrule
\sysname Stage 2 & \underline{0.0476} & \textbf{0.0158} & \underline{0.0581} & \underline{0.0423} & \underline{0.5968} & \underline{0.5313} & 0.9442 \\
\sysname Stage 3 & \textbf{0.0390} & 0.0281 & \textbf{0.0499} & \textbf{0.0218} & \textbf{0.7201} & \textbf{0.5795} & 0.9258 \\
\bottomrule
\end{tabular}
\label{tab:replay_fidelity}
\end{table}

Table~\ref{tab:replay_fidelity} shows that our Stage 3 student and Stage 2 teacher rank first and second among all methods in most DINOv2 and SSIM metrics for both self-consistency and GT reference.
Interestingly, while the student exhibits a higher GT-referenced distance on initial visits compared to the teacher ($0.0281$ vs.\ $0.0158$), it achieves superior fidelity on return visits ($0.0499$ vs.\ $0.0581$).
Conversely, baselines like HY-World 1.5 and DreamX-AR score higher on Local Memory (LM), and the teacher has higher LM than the student.
This occurs because LM prioritizes adjacent-frame temporal smoothness, which can be artificially elevated by static or over-smoothed predictions that fail to preserve long-range scene identity.

\textbf{Memory Retention during Rollout.}
We evaluate scene retention by tracking the degradation from first-visit to return-visit GT-referenced distances.
For example, as shown in Table~\ref{tab:replay_fidelity}, HY-World and our student model exhibit similar first-visit distances (both approximately $0.028$), but HY-World's return-visit error surges to $0.1867$, whereas our student's error increases marginally to $0.0499$.
This markedly lower error accumulation ($\Delta = 0.0218$ vs.\ $0.1593$) of our student indicates better retention of scene content across the evaluated revisit intervals.
These comparisons assess scene retention of the models and distinguish it from aggregate visual quality.

\subsection{Complementary Metrics for Revisit Fidelity}
\label{app:metric_robustness}

Table~\ref{tab:metric_robustness} extends the main comparison in Table~\ref{tab:main} to feature, perceptual, and pixel-level measures. Outbound-return consistency evaluates self-consistency and retention of generated content, while GT-referenced fidelity evaluates strict compliance with the reference scene.
Across all six measures, our \sysname Stage 3 and Stage 2 models consistently secure the top two ranks among all evaluated methods, validating the robust memory retention enabled by our framework.

\begin{table*}[t]
\centering
\footnotesize
\setlength{\tabcolsep}{5.0pt}
\caption{\textbf{Comprehensive Revisit Fidelity Metrics.} Scores are averaged over the four outbound-return pairs per clip across all $120$ benchmark clips. Outbound-return consistency compares each generated return view with its generated outbound view using DINOv2 feature distance ($\bar d$), SSIM, and LPIPS. GT-referenced fidelity compares generated return views with the GT references using PSNR, SSIM, and LPIPS. The VAE-decoded GT row gives the reference level for self-consistency. Bold marks the best generated result and underline denotes the second best.}
\begin{tabular}{l ccc ccc}
\toprule
& \multicolumn{3}{c}{Outbound-Return Consistency} & \multicolumn{3}{c}{GT-Referenced Fidelity} \\
\cmidrule(lr){2-4}\cmidrule(lr){5-7}
Method & DINO $\bar d\downarrow$ & SSIM $\uparrow$ & LPIPS $\downarrow$ & PSNR $\uparrow$ & SSIM $\uparrow$ & LPIPS $\downarrow$ \\
\midrule
GT (VAE-decoded) & 0.0018 & 0.9470 & 0.0221 & -- & -- & -- \\
\midrule
Matrix-Game 3.0 & 0.1775 & 0.3759 & 0.4478 & 11.15 & 0.3748 & 0.4629 \\
HY-World 1.5 & 0.1437 & 0.3576 & 0.4687 & 10.84 & 0.3526 & 0.4784 \\
DreamX-AR & 0.1232 & 0.3805 & 0.4619 & 11.66 & 0.3781 & 0.4699 \\
LingBot-Fast & 0.1420 & 0.3431 & 0.4775 & 10.59 & 0.3409 & 0.4911 \\
SANA-WM & 0.0800 & 0.4704 & 0.3805 & 13.52 & 0.4360 & 0.4272 \\
\midrule
\sysname{} Stage 2 & \underline{0.0476} & \underline{0.5968} & \underline{0.1932} & \underline{16.69} & \underline{0.5313} & \underline{0.2261} \\
\sysname{} Stage 3 & \textbf{0.0390} & \textbf{0.7201} & \textbf{0.1278} & \textbf{18.31} & \textbf{0.5795} & \textbf{0.1822} \\
\bottomrule
\end{tabular}
\label{tab:metric_robustness}
\end{table*}

\subsection{Paired Statistical Comparisons}
\label{app:paired_results}

To rigorously assess performance margins, all pairwise comparisons employ clip-level paired outcomes evaluated via exact two-sided sign tests.
Compared to SANA-WM, our Stage 3 student model achieves lower DINOv2 revisit distance on $111/120$ clips ($p = 1.7\times10^{-23}$) and higher revisit PSNR on $112/120$ clips ($p = 1.4\times10^{-24}$).
Similarly, the Stage 2 teacher model outperforms SANA-WM in DINOv2 distance on $108/120$ clips ($p = 1.8\times10^{-20}$).
On $96/120$ clips, the student model simultaneously surpasses every external baseline in both memory similarity and revisit PSNR.
Overall, our student model closes the DINOv2 feature distance gap between SANA-WM and the ground-truth limit ($0.0018$) by approximately $52\%$.

For the seven-dimensional VBench visual quality, paired sign tests favor Matrix-Game 3.0 and HY-World 1.5 over our student, while showing no statistically significant differences against DreamX-AR and LingBot-Fast. These statistical evaluations document the trade-off between peak perceptual sharpness and long-horizon memory retention.

\subsection{Camera-Sweep Analysis}
\label{app:difficulty_details}

Clips are post-hoc partitioned into three equal tiers ($40$ clips each) in Table~\ref{tab:difficulty} based on the total yaw sweep normalized by the horizontal field of view.
This stratification measures camera rotation severity without relying on model outputs.
Against SANA-WM, the student model achieves lower DINOv2 distances on $37/40$, $36/40$, and $38/40$ clips in tiers T1, T2, and T3, respectively ($p = 1.95\times10^{-8}$, $1.86\times10^{-7}$, and $1.49\times10^{-9}$).
The largest performance margin occurs in the most challenging high-sweep tier (T3).
In T3, our student achieves a rotational camera error (RotErr) of $1.288^\circ$, outperforming the teacher ($2.241^\circ$), SANA-WM ($4.316^\circ$), and remaining baselines ($11.075^\circ$-$15.477^\circ$).

\subsection{Teacher-Student Comparisons}
\label{app:distillation_details}

With self-forcing distillation, the Stage 3 student improves over the Stage 2 teacher on revisit PSNR in $91/120$ clips ($p=1.2\times10^{-8}$) and on revisit LPIPS in $86/120$ ($p=2.3\times10^{-6}$). Its lower mean DINOv2 revisit distance is also supported by $72/120$ wins ($p=0.0353$). First-visit PSNR and all four camera metrics yield $p$-values between $0.14$ and $0.36$, indicating comparable trajectory control between teacher and student. 
As shown in Table \ref{tab:metric_robustness}, revisit SSIM also improves descriptively from $0.5313$ to $0.5795$, while Local Memory (LM) decreases modestly from $0.9442$ to $0.9258$, mirroring the perceptual quality trade-offs observed under VBench.

\section{Inference Efficiency}
\label{app:inference}

Our 4-step \sysname student achieves $20.6$\,FPS at $1280\times704$ on a single NVIDIA B200 GPU with online memory updates enabled, as shown in Figure~\ref{fig:teaser-speed}. The teacher uses $50$ denoising steps per block and the student uses $4$, a $12.5\times$ reduction in step count. We report measured speedup separately from this ratio. This section describes how the 4-step student reaches real-time generation on a single GPU.
All runtime efficiency optimizations do not change the model architecture and preserve the numerical fidelity: the student weights, its 4-step schedule, the latent-frame context window size, and the TTT memory with online updates are unchanged.
Appendix~\ref{app:eff_protocol} defines the workload and settings, Appendix~\ref{app:eff_methods} introduces the optimizations, Appendix~\ref{app:eff_results} demonstrates their measured effect, and Appendix~\ref{app:eff_fidelity} tests that the optimization remains numerically equivalent to the baseline.
Appendix~\ref{app:eff_baselines} compares inference cost across the baselines of Table~\ref{tab:main} on the same hardware.

\subsection{Workload and Measurement Settings}
\label{app:eff_protocol}

\textbf{Per-Block Workload.}
The \sysname student inherits the Wan2.2-TI2V-5B backbone~\citep{wan2025}: $30$ DiT blocks of width $3072$ with $24$ attention heads, and a causal 3D VAE with $16\times$ spatial and $4\times$ temporal compression.
At $1280\times704$, a latent frame contains $880$ tokens.
A generation block holds $B=3$ latent frames ($2{,}640$ tokens) and decodes to $12$ video frames, i.e., $0.75$\,s at $16$\,FPS.
For every block, the implementation performs four steps:
(i) it encodes the clean context, i.e., the persistent sink and the most recent clean blocks, at most $18$ latent frames or $15{,}840$ tokens, under the block-causal mask and caches the resulting keys and values,
(ii) it runs the four denoising forwards of the target block, each attending to the cached context and to the block itself,
(iii) it writes the block evicted from the window into the TTT memory (Eq.~\ref{eq:rls}) and reads the memory at the target pose (Eq.~\ref{eq:read}), and
(iv) it decodes the finished block with the causal VAE decoder, carrying the decoder's temporal convolution cache from block to block.
The context is re-encoded at every block rather than extended incrementally: when the window slides, each context block loses its oldest predecessor, so its hidden states change.
Once the window is full, the per-block cost is therefore constant and does not depend on the rollout length.
The fixed-size fast-weight and RLS states likewise keep memory storage constant (Sec.~\ref{sec:stage2}).

\textbf{Metrics.}
\emph{End-to-end latency} is the wall-clock time from the inputs on disk (camera trajectory, first frame, and text prompt) to the last generated frame as 8-bit RGB in host memory.
It includes first-frame VAE encoding, text encoding, camera embedding, sampling, TTT memory updates, decoding, and the device-to-host copy, and it excludes one-time model loading and video-container writing.
We report it directly, as frames divided by latency (FPS), and as wall-clock seconds per second of generated video.
\emph{Peak memory} is the device memory in use, polled through NVML every $50$--$100$\,ms during the timed rollouts.

\textbf{Settings.}
Every configuration runs in its own warm process on NVIDIA B200 GPUs ($180$\,GB HBM3e): one discarded warm-up rollout absorbs compilation, kernel autotuning, and allocator growth, and we report the median of three timed rollouts.
Runs execute one at a time on an $8$-GPU node.
We time two lengths: L5 ($81$ frames, $5.06$\,s), the training horizon, and L30 ($465$ frames, $29.06$\,s), where the window has been full for most of the rollout.
Our model runs in default bf16 with PyTorch~2.15 (development build), CUDA~12.8, and cuDNN~9.18.

\subsection{Optimizations}
\label{app:eff_methods}

We group the optimizations by the source of the saving.
Table~\ref{tab:eff_ladder} applies them cumulatively, in the order in which we added them.

\textbf{Reusing Features across Denoising Steps and Blocks.}
The keys and values of the clean context are computed once per block and reused by all four denoising steps. This KV cache is part of the starting point.
We extend reuse in three places.
First, the TTT memory state $\mathcal{M}_b$ and the target pose key $\kk_b$ do not change during the four steps, so the memory feature $\mathbf{m}_b$ of Eq.~\ref{eq:read} is computed once per block instead of once per step, and the output is bit-identical.
Second, context keys are cached before rotary position embedding (RoPE)~\citep{roformer2024} because their positions shift as the window slides, and we rotate them once per block rather than at every step.
Third, under the block-causal mask, the keys and values of a context block depend only on that block and the context blocks before it.
Blocks that form an unchanged prefix of the previous block's context therefore have unchanged keys and values and are not recomputed: all earlier context blocks while the window is still filling, and the sink afterwards.

\textbf{Compilation and Attention Kernels.}
We compile the per-layer bodies of the denoising forward and of the context encoder with \texttt{torch.compile}~\citep{torchcompile2024}, which fuses the elementwise chains around attention (adaptive layer-norm modulation, gating, normalization) and keeps their intermediates in FP32.
Self-attention in both passes uses FlashAttention-4~\citep{flashattention4}, which targets the Blackwell architecture, in place of FlexAttention~\citep{flexattention2025}. The block-causal context attention is issued as one dense call per query block over its visible key prefix, and both calls are registered as custom operators so that they remain inside the compiled graphs.
The reference RoPE casts queries and keys to FP64 and multiplies them by complex128 frequencies, and inside a compiled graph this runs as separate unfused kernels with FP64 temporaries.
We apply the same rotation in real FP32 arithmetic, whose relative error (about $6\times10^{-8}$) is far below the bf16 resolution of the rotated tensors, so the rotation fuses with the preceding normalization.

\textbf{VAE Decoding.}
We enable cuDNN~9~\citep{cudnn2014} convolutions, store decoder weights and activations in channels-last (NDHWC) layout, and let cuDNN select its fastest engine for each convolution shape by benchmarking all candidates, restricting the compiler's autotuning to cuDNN convolutions.
The causal convolutions of the reference decoder also materialize a zero-padded copy of every input. We pass the spatial padding to the convolution itself and keep only the temporal padding, which comes from the causal cache.
Decoding block by block with the causal cache performs the same convolutions as decoding the full latent sequence at once.

\textbf{Host-Side Overhead.}
Two host-side operations left the GPU idle between kernels.
The RoPE frequency tables were rebuilt on the CPU and copied from pageable host memory at every denoising step, five $16$\,MB transfers per block. We build them once per window shape and keep them on the device.
The per-block Gaussian noise is drawn from a CPU generator for reproducibility, and each draw with its synchronous copy stalled the stream. A background thread now draws the noise of upcoming blocks from the same generator in the same order into pinned memory, so the transfer is asynchronous.
Both changes leave the output bit-identical.

\textbf{Multiple GPUs.}
In the \rtdeploy, a second GPU decodes block $b$ while the first samples block $b{+}1$. The decoder state stays on the second GPU, so the decoding is the same as in the serial schedule.
For higher single-stream throughput, we additionally split the tokens of each forward across GPUs with Ulysses sequence parallelism~\citep{ulysses2023}, which exchanges them through all-to-all communication before and after attention, replay the per-layer loops as CUDA graphs~\citep{cudagraphs} to remove kernel-launch overhead, and split the decoder layer-wise into a two-stage pipeline over two further GPUs.

\subsection{Results}
\label{app:eff_results}

\begin{table}[t]
\centering
\begin{minipage}[t]{0.58\linewidth}
\centering
\captionof{table}{\textbf{Cumulative Effect of the Optimizations on one B200.} Each row is a separate warm process, and values are medians of three generations of a $30$\,s video ($465$ frames at $16$\,FPS). E2E is end-to-end latency, FPS is frames per second of latency, and Mem.\ is peak GPU memory in GB.}
\label{tab:eff_ladder}
\vspace{-2mm}
\setlength{\tabcolsep}{3pt}
\resizebox{0.99\textwidth}{!}{
\begin{tabular}{r p{6.0cm} ccc}
\toprule
\# & Change (cumulative) & E2E (s) & FPS & Mem. \\
\midrule
1 & Baseline & 70.48 & 6.6 & 72.1 \\
\midrule
2 & + \texttt{torch.compile} & 63.55 & 7.3 & 71.6 \\
3 & + cuDNN~9, channels-last VAE decoder & 37.23 & 12.5 & 52.8 \\
4 & + FlashAttention-4 for self-attention & 30.00 & 15.5 & 52.8 \\
5 & + memory read once per block & 29.55 & 15.7 & 52.8 \\
6 & + RoPE of cached keys once per block & 29.05 & 16.0 & 52.8 \\
7 & + context prefix reuse & 27.56 & 16.9 & 54.9 \\
8 & + shape-aware conv.\ engines & 24.25 & 19.2 & 54.4 \\
9 & + prefetch Gaussian noise& 23.83 & 19.5 & 53.5 \\
10 & + FP32 Rotation & 22.59 & 20.6 & 54.3 \\
\bottomrule
\end{tabular}}
\end{minipage}\hfill
\begin{minipage}[t]{0.4\linewidth}
\centering
\captionof{figure}{\textbf{Time per Steady-State Block on one GPU}, start (\#1) vs.\ final (\#10): CUDA-event timers over the $32$ full-window blocks of a $30$\,s rollout.}
\label{fig:eff_breakdown}
\vspace{-1mm}
\includegraphics[width=\linewidth]{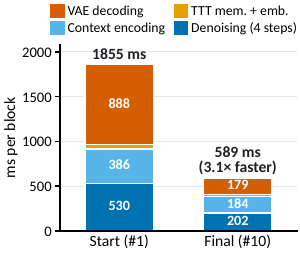}
\end{minipage}
\vspace{-3mm}
\end{table}

Table~\ref{tab:eff_ladder} applies the optimizations cumulatively to a $30$\,s rollout on one GPU.
End-to-end latency falls from $70.48$ to $22.59$\,s ($3.12\times$).
Both stages contribute: sampling time per block falls from $943$ to $391$\,ms and decoding time from $890$ to $182$\,ms.

Figure~\ref{fig:eff_breakdown} splits a steady-state block into its phases.
In the starting point, decoding takes almost half of the block ($888$ of $1855$\,ms).
After optimization, the four denoising forwards and the context encoding account for about two thirds of the block ($386$ of $589$\,ms), and decoding for $179$\,ms.
One context encoding costs about as much as all four denoising forwards together ($184$ vs.\ $202$\,ms): with the sink reused, it still processes five times as many query tokens as a single forward ($13{,}200$ vs.\ $2{,}640$).
The remaining $24$\,ms include the TTT memory write, the timestep and camera embeddings, scheduling, and host-side work.

\subsection{Numerical Fidelity}
\label{app:eff_fidelity}

\begin{table}[t]
\centering
\footnotesize
\setlength{\tabcolsep}{5pt}
\caption{\textbf{Numerical Fidelity of the Accelerated Implementation.} Operator-level errors use the captured inputs of block~$5$ of a held-out clip and an FP32 copy of the model with TF32 disabled, and the decoder decodes the same $21$ latent frames. Rollout rows compare full $81$-frame bf16 rollouts with FP32 rollouts from the same inputs and seed. }
\vspace{-2mm}
\begin{tabular}{c c cc}
\toprule
Test & Measure & Reference bf16 & Accelerated \\
\midrule
Denoising forward & rel.\ RMS error to FP32 $\downarrow$ & $1.54\times10^{-2}$ & $8.46\times10^{-3}$ \\
Context encoding & rel.\ RMS error to FP32 $\downarrow$ & $1.28\times10^{-2}$ & $8.25\times10^{-3}$ \\
\multirow{2}{*}{VAE decoder} & rel.\ RMS error to FP32 $\downarrow$ & $7.17\times10^{-3}$ & $5.95\times10^{-3}$ \\
 & PSNR to FP32 (dB) $\uparrow$ & 53.82 & 55.44 \\
\midrule
\multirow{2}{*}{Rollout} & PSNR to FP32 rollout (dB) $\uparrow$ & 27.40 & 29.61 \\
 & SSIM to FP32 rollout $\uparrow$ & 0.9034 & 0.9257 \\
\bottomrule
\end{tabular}
\label{tab:eff_fidelity}
\end{table}

Most of the proposed optimizations in Table~\ref{tab:eff_ladder} produce bit-identical outputs.
The remaining optimizations reorder floating-point operations (row 3, 8), select different kernels (row 4), change the precision of the RoPE rotation (row 10), so their outputs differ from the reference at the level of floating-point rounding.
Comparing generated videos against the reference would not measure this difference: the autoregressive sampler amplifies rounding differences into a different but equally valid rollout.
We therefore test at two levels (Table~\ref{tab:eff_fidelity}).

\textbf{Operator Level.}
For each implementation, we capture the inputs of one denoising forward and one context encoding at a steady-state block (block~$5$) of a held-out clip, and compare its bf16 outputs with those of an FP32 copy of the model evaluated on FP32 copies of the same inputs, with TF32 disabled.
The accelerated implementation passes if its relative RMS error to the FP32 output is at most $1.05$ times that of the reference. This criterion was fixed before measurement.
The decoder is tested analogously on the same latents: the eager bf16 decoder, the accelerated decoder, and an FP32 decoder decode the $21$ latent frames of one clip.
The accelerated implementation is closer to the FP32 result than the reference in all three cases, mainly because compiled kernels keep fused intermediates in FP32 where the reference rounds to bf16 after every operation.

\textbf{Rollout Level.}
We generate complete $81$-frame rollouts with an FP32 model and measure how closely each bf16 implementation reproduces them.
The accelerated rollout is closer to the FP32 rollout than the reference bf16 rollout on $7$ of $8$ clips.
For scale, the reference rollout with a different noise seed reaches only $17.90$\,dB against the FP32 rollout, so the differences between bf16 implementations are small compared with sampling variability.
The sequence-parallel sampler with CUDA graphs, tested with three GPUs, and the layer-pipelined decoder pass the same operator-level criterion (error ratios $0.56$ for the denoising forward, $0.62$ for the context encoding, and a decoder error of $5.96\times10^{-3}$ against $7.19\times10^{-3}$ for the eager decoder).
The replay-benchmark videos in Sec.~\ref{sec:experiments} were generated with the evaluation implementation and were not regenerated with the accelerated one. The tests above address their equivalence directly.

\subsection{Efficiency Comparison with Other Video World Models}
\label{app:eff_baselines}

\begin{table}[t]
\centering
\small
\setlength{\tabcolsep}{3.6pt}
\caption{\textbf{Inference Cost Comparison on NVIDIA B200 GPUs.} Metrics are the median over three generations, run one at a time. L30 is the $30$\,s video setting. Mem.\ is peak memory in GB per GPU. For multi-GPU rows, we use maximum peak memory over GPUs.}
\vspace{-2mm}
\begin{tabular}{l c c rr rrr}
\toprule
 & & & \multicolumn{2}{c}{E2E latency (s)} & \multicolumn{3}{c}{L30} \\
\cmidrule(lr){4-5}\cmidrule(lr){6-8}
Method & Steps & GPUs & 5s & 30s & s\,/\,video s & FPS &  Peak Mem (GB) \\
\midrule
\rowcolor{gray!10} \multicolumn{8}{c}{One GPU} \\
Matrix-Game 3.0 & 3 & 1 & 12.23 & 69.73 & 2.24 & 7.1 &  76 \\
HY-World 1.5 & 4 & 1 & 7.96 & 107.04 & 3.59 & 4.5 & 110 \\
DreamX-AR & 4 & 1 & 7.23 & 42.22 & 1.45 & 11.0 &  76 \\
LingBot-Fast & 4 & 1 & 44.56 & 292.54 & 10.07 & 1.6 & 144 \\
SANA-WM & 4 & 1 & 6.69 & 32.75 & 1.13 & 14.2 &  48 \\
\sysname{} Stage 3 & 4 & 1 & 3.63 & 22.59 & 0.78 & 20.6 & 54 \\
\midrule
\rowcolor{gray!10} \multicolumn{8}{c}{Multiple GPUs} \\
Matrix-Game 3.0 & 3 & 8 & 6.50 & 41.95 & 1.35 & 11.8 &  61 \\
HY-World 1.5 & 4 & 8 & 2.30 & 24.16 & 0.81 & 19.7 &  70 \\
LingBot-Fast & 4 & 8 & 16.72 & 100.32 & 3.45 & 4.6 &  107 \\
\sysname{} Stage 3 & 4 & 8 & 1.79 & 8.28 & 0.28 & 56.2 & 47 \\
\bottomrule
\end{tabular}
\label{tab:eff_cross}
\end{table}

Table~\ref{tab:eff_cross} gives a more detailed measurement of all the baselines in Table~\ref{tab:main} with both the single-GPU and multiple-GPU settings.
Each baseline runs its official inference code with the configuration used for its quality results in Table~\ref{tab:main}.
Because the models produce different numbers of frames per chunk, their clips differ slightly in length (between $464$ and $497$ frames for L30), while seconds per second of video normalizes for this.
On one GPU, our student is the only model that generates a $30$\,s rollout faster than real time. SANA-WM reports peak memory from the PyTorch allocator instead of the device, and runs its $4$-step generator followed by a $3$-step refiner. 
With multiple GPUs, HY-World~1.5 also becomes faster than real time, while our model remains the most efficient. Multi-GPU baselines use their official parallel modes, which is not supported by DreamX-AR and SANA-WM.

\section{Limitations}
\label{app:limitation}

Although \sysname demonstrates robust persistent scene memory and interactive generation in real-time, several limitations remain to be addressed in future work.

Our framework relies on 3D foundation models, such as VGGT-$\Omega$ \citep{vggtomega2026}, to extract 6-DoF camera trajectories from video corpora for conditioning. However, off-the-shelf pose estimators occasionally exhibit tracking noise or subtle drift, particularly in challenging visual environments with weak textures or rapid camera maneuvers. Since our test-time memory consolidation performs pose-keyed retrieval, such inaccuracies in the camera trajectory can degrade the alignment between queries and fast-weight states. Over extended generation horizons, such pose estimation errors may also accumulate and lead to suboptimal memory recall during viewpoint revisits. Addressing this limitation necessitates training on higher-quality camera trajectory datasets or incorporating more robust camera estimation pipelines.

Additionally, while our pose-indexed test-time training memory improves temporal stability and viewpoint revisit consistency over extended rollouts, maintaining pristine visual fidelity and generating well beyond the training context length remain open challenges in video world modeling. As autoregressive rollouts proceed significantly past the temporal window seen during training, video frames can occasionally experience gradual visual attenuation or fine-grained texture degradation due to compounding errors. Achieving photorealistic high-resolution generation and unbounded temporal extrapolation demands further advances in scalable generative architectures, higher-capacity latent spaces, and dynamic context management strategies.

Despite these limitations, \sysname provides a viable and efficient path toward persistent memory in interactive video world models, and we hope our TTT memory formulation inspires future explorations in real-time long-horizon scene generation.

\end{document}